%% file: main.tex
\documentclass[Afour,sageh,times]{sagej}

\usepackage{moreverb,url}

\usepackage[colorlinks,bookmarksopen,bookmarksnumbered,citecolor=red,urlcolor=red]{hyperref}
\usepackage[dvipsnames]{xcolor}
\usepackage{soul}
\usepackage[normalem]{ulem}
\newcommand\BibTeX{{\rmfamily B\kern-.05em \textsc{i\kern-.025em b}\kern-.08em
T\kern-.1667em\lower.7ex\hbox{E}\kern-.125emX}}

\def\volumeyear{2020}

\include{defs}

\hypersetup{draft}

\begin{document}

\runninghead{Ibarz, Tan, Finn, Kalakrishnan, Pastor, Levine}

\title{How to Train Your Robot with Deep Reinforcement Learning -- Lessons We've Learned}

\author{Julian Ibarz\affilnum{1}, Jie Tan\affilnum{1}, Chelsea Finn\affilnum{1,3}, Mrinal Kalakrishnan\affilnum{2}, Peter Pastor\affilnum{2}, Sergey Levine\affilnum{1,4}}

\affiliation{\affilnum{1}Robotics at Google\\
\affilnum{2}Everyday Robots, X, The Moonshot Factory\\
\affilnum{3}Stanford University\\
\affilnum{4}University of California Berkeley}

\corrauth{Julian Ibarz}

\email{julianibarz@google.com}

\begin{abstract}
Deep reinforcement learning (RL) has emerged as a promising approach for autonomously acquiring complex behaviors from low level sensor observations.
Although a large portion of deep RL research has focused on applications in video games and simulated control, which does not connect with the constraints of learning in real environments, deep RL has also demonstrated promise in enabling physical robots to learn complex skills in the real world.
At the same time, real world robotics provides an appealing domain for evaluating such algorithms, as it connects directly to how humans learn -- as an embodied agent in the real world.
Learning to perceive and move in the real world presents numerous challenges, some of which are easier to address than others, and some of which are often not considered in RL research that focuses only on simulated domains.
In this review article, we present a number of case studies involving robotic deep RL. Building off of these case studies, we discuss commonly perceived challenges in deep RL and how they have been addressed in these works. We also provide an overview of other outstanding challenges, many of which are unique to the real-world robotics setting and are not often the focus of mainstream RL research.
Our goal is to provide a resource both for roboticists and machine learning researchers who are interested in furthering the progress of deep RL in the real world.
\end{abstract}


\maketitle

\section{Introduction}
\label{sec:introduction}

Robotic learning lies at the intersection of machine learning and robotics. 
From the perspective of a machine learning researcher interested in studying intelligence, robotics is an appealing medium to study as it provides a lens into the constraints that humans and animals encounter when learning, uncovering aspects of intelligence that might not otherwise be apparent to study when we restrict ourselves to simulated environments.
For example, robots receive streams of raw sensory observations as a consequence of their actions, and cannot practically obtain large amounts of detailed supervision beyond observing these sensor readings.
This makes for a challenging but highly realistic learning problem.
Further, unlike agents in video games, robots do not readily receive a score or reward function
that is shaped for their needs, and instead need to develop their own internal representation of progress towards goals.
From the perspective of robotics research, using learning-based techniques is appealing because it can enable robots to move towards less structured environments, to handle unknown objects, and to learn a state representation suitable for multiple tasks.

Despite being an interesting medium, there is a significant barrier for a machine learning researcher to enter robotics and vice-versa. Beyond the cost of a robot, there are many design choices in choosing how to set-up the algorithm and the robot. For example, reinforcement learning (RL) algorithms require learning from experience that the robot autonomously collects itself, opening up many choices in how the learning is initialized, how to prevent unsafe behavior, and how to define the goal or reward. \new{Likewise, machine learning and RL algorithms also provide a number of important design choices and hyperparameters that can be tricky to select.}

Motivated by these challenges for the researchers in the respective fields, our goal in this article is to provide a high-level overview of how deep RL can be approached in a robotics context, summarize the ways in which key challenges in RL have been addressed in some of our own previous work, and provide a perspective on major challenges that remain to be solved, many of which are not yet the subject of active research in the RL community.

There have been high-quality survey articles about applying machine learning to robotics.~\cite{deisenroth2013survey} focused on policy search techniques for robotics, whereas~\cite{kober2013reinforcement} focused on RL. More recently,~\cite{manipulation_review} reviewed the learning algorithms for manipulation tasks. ~\cite{deep_learning_for_robotics} identified current areas of research in deep learning that were relevant to robotics, and described a few challenges in applying deep learning techniques to robotics.

Instead of writing another comprehensive literature review, we first center our discussion around three case studies from our own prior work. We then provide an in-depth discussion of a few topics that we consider especially important given our experience. This article naturally includes numerous opinions. When sharing our opinions, we do our best to ground our recommendations in empirical evidence, while also discussing alternative options.
We hope that, by documenting these experiences and our practices, we can provide a useful resource both for roboticists interested in using deep RL and for machine learning researchers interested in working with robots.

\section{Background}
\new{
In this section, we provide a brief, informal introduction to RL, by contrasting it with classical techniques of programming robot behavior.
A robotics problem is formalized by defining a \emph{state} and \emph{action} space, and the \emph{dynamics} which describe how actions influence the state of the system. The state space includes internal states of the robot as well as the state of the world that is intended to be controlled. Quite often, the state is not directly observable -- instead, the robot is equipped with sensors, which provide \emph{observations} that can be used to infer the state. The goal may be defined either as a target state to be achieved, or as a \emph{reward function} to be maximized. We want to find a controller, (known as a \emph{policy} in RL parlance), that maps states to actions in a way that maximizes the reward when executed.
}

\new{
If the states can be directly or indirectly observed, and a model of the system dynamics is known, the problem can be solved with classical methods such as planning or optimal control. These methods use the knowledge of the dynamics model to search for sequences of actions that when applied from the start state, take the system to the desired goal state or maximize the achieved reward.
However, if the dynamics model is unknown, the problem falls into the realm of RL~\citep{sutton2018reinforcement}. In the paradigm of RL, samples of state-action sequences (\emph{trajectories}) are required in order to learn how to control the robot and maximize the reward. In \emph{model-based RL}, the samples are used to learn a dynamics model of the environment, which in turn is used in a planning or optimal control algorithm to produce a policy or the sequence of controls. In \emph{model-free RL}, the dynamics are not explicitly modeled, but instead the optimal policy or value function is learned directly by interaction with the environment. Both model-based and model-free RL have their own strengths and weaknesses, and the choice of algorithm depends heavily on the properties required. These considerations are discussed further in Sections~\ref{sec:case_studies} and~\ref{sec:tameable_challenges}.
}

\section{Case Studies in Robotic Deep RL}
\label{sec:case_studies}

In this section, we present a few case studies of applications of deep RL to various robotic tasks that we have studied. The applications span manipulation, grasping, and legged locomotion. The sensory inputs used range from low-dimensional proprioceptive state information to high-dimensional camera pixels, and the action spaces include both continuous and discrete actions.

By consolidating our experiences from those case studies, we seek to derive a common understanding of the kinds of robotic tasks that are tractable to solve with deep RL today. Using these case studies as a backdrop, we point readers to outstanding challenges that remain to be solved and are commonly encountered in Section~\ref{sec:tameable_challenges}. \old{Whenever possible, we share some strategies that we found useful to mitigate these challenges.}

\subsection{Learning manipulation skills}

Reinforcement learning of individual robotic skills has a long history~\citep{pma-reps-10,ins-minds-02,ps-rlmsp-08,kkgb-rldcs-12,dnkp-lsmt-13,kober2013reinforcement,mkgp-lsmpd-14}.
Deep RL provides some appealing capabilities in this regard: deep neural network policies can alleviate the need to manually design policy classes, provide a moderate amount of generalization to variable initial conditions and, perhaps most importantly, \new{allow for end-to-end joint training for both perception and control, learning to directly map high-dimensional sensory inputs, such as images, to control outputs. Of course, such end-to-end training itself presents a number of challenges, which we will also discuss.}\old{make it practical to handle high-dimensional sensory inputs, such as images.} We discuss a few case studies on single-task deep robotic learning with a variety of different methods, including model-based and model-free algorithms, and with different starting assumptions.

\subsubsection{Guided policy search.}

\begin{figure}[th]
\begin{center}
\includegraphics[width=0.6\columnwidth]{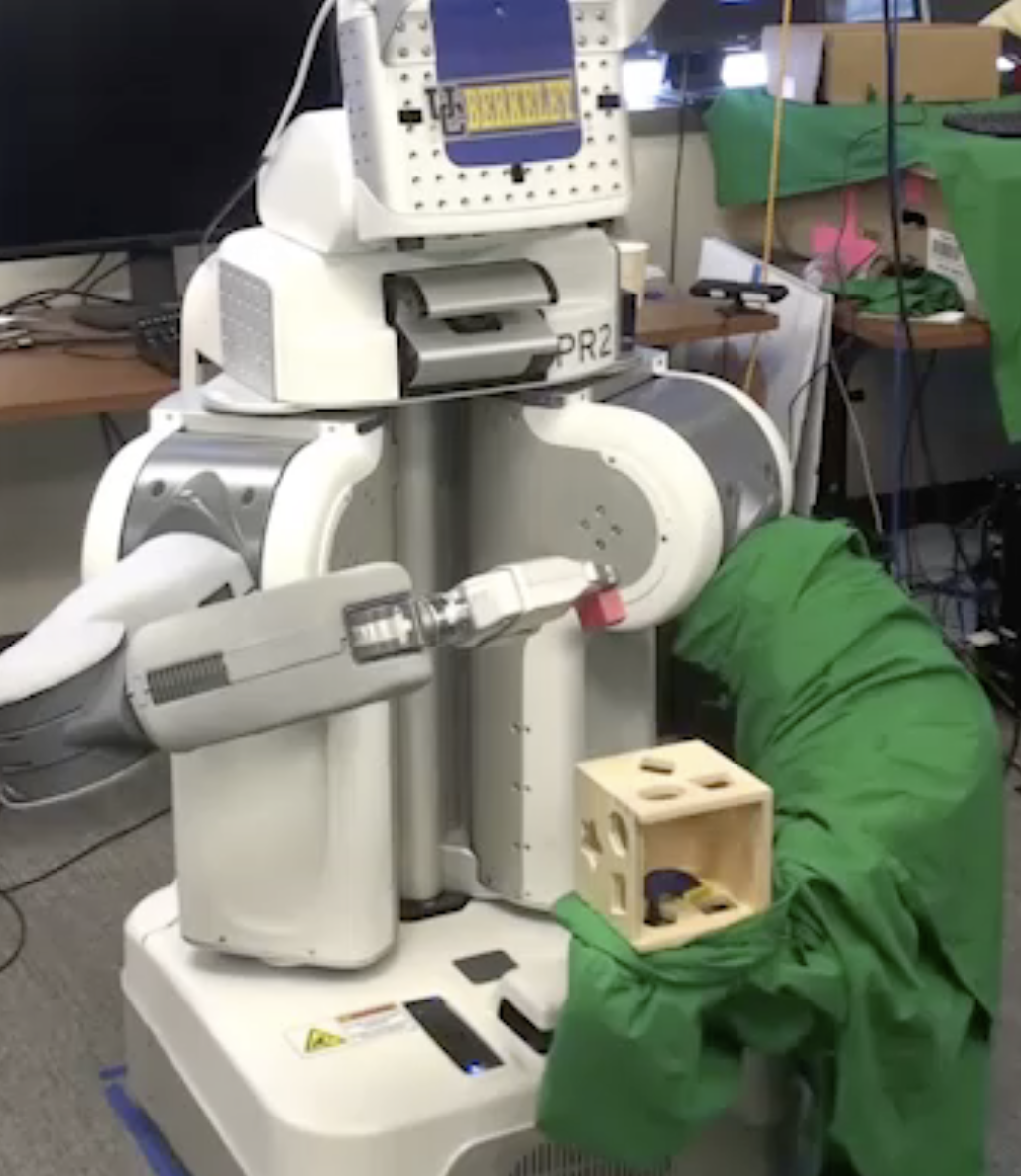}
\caption{A PR2 learns to place a red trapezoid block into a shape-sorting cube. With~\citet{end2end}, it learns local policies for each initial position of the cube, which can be reset automatically using the robot's left arm. The local policies are distilled into a global policy that takes images as input, rather than the cube's location.
}
\label{fig:shape_sorting_cube}
\end{center}
\end{figure}

Guided policy search methods~\citep{end2end}
were among the first deep RL methods that could be tractably applied to learn individual neural network skills for image-based manipulation tasks. The basic principle behind these methods is that the neural network policy is ``guided'' by another RL method, typically a model-based RL algorithm. The neural network policy is referred to as a \emph{global policy}, and is trained to perform the task successfully from raw sensory observations and under moderate variability in the initial conditions. For example, as shown in Figure~\ref{fig:shape_sorting_cube}, the global policy might be required to put the red shape into the shape sorting cube at different positions. \new{This requires the policy to \emph{implicitly} determine the position of the hole. However, this is not supervised directly, but instead the perception mechanism is learned end-to-end together with control.}\old{, using vision to determine the position of the hole.} Supervision is provided from multiple individual model-based learners that learn separate \emph{local policies} to insert the shape into the hole at a specific position. In the case of the experiment illustrated in Figure~\ref{fig:shape_sorting_cube}, nine local policies were trained for nine different cube positions, and a single global policy was then trained to perform the task from images. Typically, the local policies do not use deep RL, and do not use image inputs. They instead use observations that reflect the low-dimensional, ``true'' state of the system, such as the position of the shape-sorting cube in the previous example, in order to learn more efficiently. 
Local policies can be trained with model-based methods such as LQR-FLM~\citep{levine2014learning, end2end},
which uses LQR with fitted time-varying linear models, or model-free techniques such as PI2~\citep{pigps,pilqr}. 

A full theoretical treatment of the guided policy search algorithm is outside the scope of this article, and we refer the reader to prior work on this topic~\citep{levine2013guided,levine2014learning,end2end}.

An important point of discussion for this article, however, is the set of assumptions underlying guided policy search methods. Typically, such methods assume that the local policies can be optimized with simple, ``shallow'' RL methods, such as LQR-FLM or PI2. 
This assumption is reasonable for robotic manipulation tasks trained in laboratory settings, but can prove difficult in \textbf{(1)} open-world environments where the low-level state of the system cannot be effectively measured and in \textbf{(2)} settings where resetting the environment poses a challenge. For example, in the experiment in Figure~\ref{fig:shape_sorting_cube}, the robot is holding the cube in its left arm during training, so that the position of the cube can be provided to the low-level policies and so that the robot can automatically reposition the cube into different positions deterministically. We discuss these challenges in more detail in Sections~\ref{subsec:robot_persistence} and~\ref{sec:input_remapping}.

Nonetheless, for learning individual robotic skills, guided policy search methods have been applied widely and to a broad range of behaviors, ranging from inserting objects into containers and putting caps on bottles~\citep{end2end}, opening doors~\citep{pigps}, and shooting hockey pucks~\citep{pilqr}. In most cases, guided policy search methods are \new{very efficient in terms of the number of samples, particularly as compared to model-free RL algorithms, since the model-based local policy learners can acquire the local solutions quickly and efficiently.}\old{exceedingly sample-efficient, since the model-based local policy learners require only a modest number of samples.} Image-based tasks can typically be learned in a few hundred trials, corresponding to 2-3 hours of real-world training, including all resets and network training time~\citep{end2end,pilqr}.

\subsubsection{Model-free skill learning.}

\begin{figure}
    \centering
    \includegraphics[width=0.99\columnwidth]{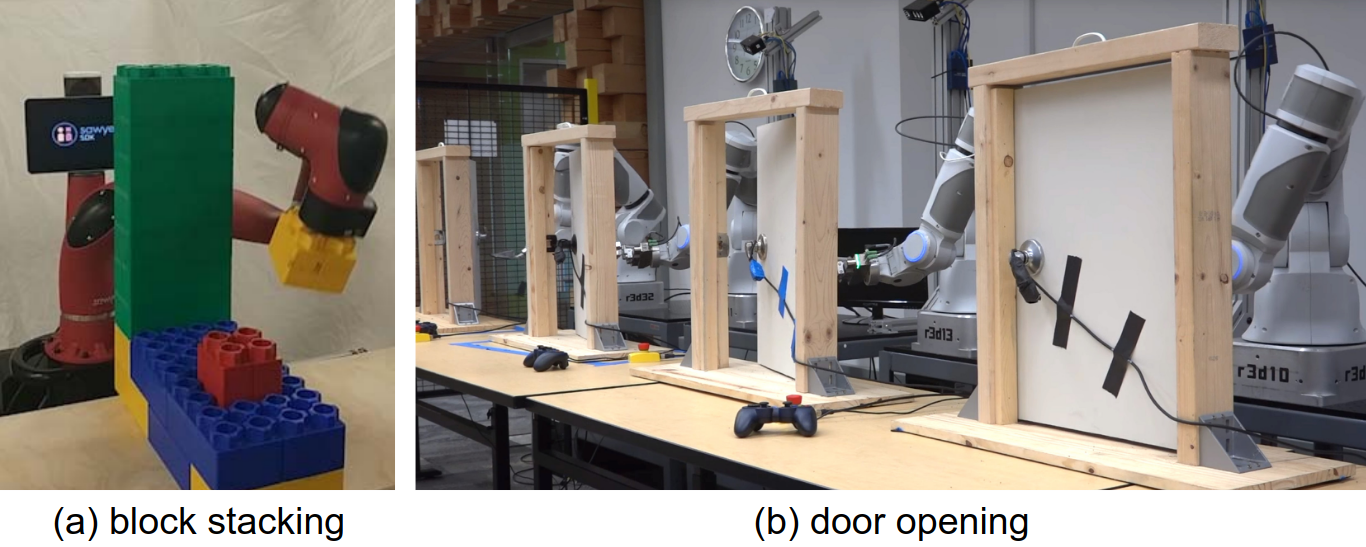}
    \vspace{-0.1in}
    \caption{Examples of model-free based algorithms learning skills in a few hours from low-dimensional state observations. (a) is learning to stack lego blocks with~\cite{haarnoja2018composable}. (b) is learning door opening with~\cite{gu2017deep}.}
    \label{fig:model_free_from_state}
    \vspace{-0.2in}
\end{figure}

Model-free RL algorithms lift some of the limitations of guided policy search, such as the need to decompose a task into multiple distinct and repeatable initial states or the need for a model-based optimizer that typically operates on a low dimensional state representation, but at the cost of a substantial increase in the required number of samples. For example, the Lego block stacking experiment reported by \citet{haarnoja2018composable}
required a little over two hours of interaction, whereas comparable Lego block stacking experiments reported by \citet{levine1501learning}
required about 10 minutes of training. The gap in training time tends to close a bit when we consider tasks with more variability: guided policy search generally requires a linear increase in the number of samples with more initial states, whereas model-free algorithms can better integrate experience from multiple initial states and goals, typically with sub-linear increase in sample requirements. As model-free methods generally do not require a lower-dimensional state for model-based trajectory optimization, they can also be applied to tasks that can only be defined on images, without an explicit representation learning phase.

Although there is a long history of model-free RL in robotics~\citep{pma-reps-10,ins-minds-02,ps-rlmsp-08,kkgb-rldcs-12,dnkp-lsmt-13,kober2013reinforcement,mkgp-lsmpd-14},
modern model-free deep RL algorithms have been used more recently for tasks such as door opening~\citep{gu2017deep}
and assembly and stacking of objects~\citep{haarnoja2018composable} with low-dimensional state observations. These methods were generally based on off-policy actor-critic designs, such as DDPG or NAF~\citep{lillicrap2015continuous,gu2016continuous},
soft Q-learning~\citep{haarnoja_sac,haarnoja2018composable},
and soft actor-critic~\citep{haarnoja2018}. An illustration of some of these tasks is shown in Figure~\ref{fig:model_free_from_state}. From our experiences, we generally found that simple manipulation tasks, such as opening doors and stacking Lego blocks, either with a single position or some variation in position, can be learned in 2-4 hours of interaction, with either torque control or end-effector position control. Incorporating demonstration data and other sources of supervision can further accelerate some of these methods~\citep{vevcerik2017leveraging,riedmiller2018learning}. \new{Section~\ref{subsec:sample_efficiency} describes other techniques to make those approaches more sample efficient.}

Although most model-free deep RL algorithms that have been applied to learn manipulation skills directly from real-world data have used off-policy algorithms based on Q-learning~\citep{haarnoja2018composable,gu2017deep} or actor-critic designs~\citep{haarnoja_sac},
on-policy policy gradient algorithms have also been used. Although standard configurations of these methods can require around 10 times the number of samples as off-policy algorithms, on-policy methods such as TRPO~\citep{schulman2015}, NPG~\citep{kakade2002natural},
and PPO~\citep{schulman2017proximal} can be tuned to only be 2-3 times less efficient than off-policy algorithms in some tasks~\citep{peng2019advantage}.
In some cases, this increased sample requirement may be justified by ease of use, better stability, and better robustness to suboptimal hyperparameter settings. On-policy policy gradient algorithms have been used to learn tasks such as peg insertion~\citep{lee2018making}, targeted throwing~\cite{ghadirzadeh2017deep},
and dexterous manipulation~\citep{zhu2018dexterous}
directly on real-world hardware, and can be further accelerated with example demonstrations~\citep{zhu2018dexterous}.

While in principle model-free deep RL algorithms should excel at learning directly from raw image observations, in practice this is a particularly difficult training regime, and good real-world results with model-free deep RL learning directly from raw image observations have only been obtained recently, with accompanying improvements in the efficiency and stability of off-policy model-free RL methods~\citep{haarnoja2018,haarnoja_sac,fujimoto2018addressing}.
The SAC algorithm can learn tasks in the real world directly from images~\citep{haarnoja2018,singh2019end}, and several other recent works have studied real-world learning from images~\citep{schoettler2019deep,schwab2019simultaneously}.

All of these experiments were conducted in relatively constrained laboratory environments, and although the learned skills use raw image observations, they generally have limited robustness to realistic visual perturbations and can only handle the specific objects on which they are trained. We discuss in Section~\ref{sec:grasping} how image-based deep RL can be scaled up to enable meaningful generalization. Furthermore, a major challenge in learning from raw image observations in the real world is the problem of reward specification: if the robot needs to learn from raw image observations, it also needs to evaluate the reward function from raw image observations, which itself can require a hand-designed perception system, partly defeating the purpose of learning from images in the first place, or otherwise require extensive instrumentation of the environment~\citep{zhu2018dexterous}. We discuss this challenge further in Section~\ref{sec:rewards}.

\subsubsection{Learning predictive models for multiple skills with visual foresight.}
\label{subsec:foresight}

Although there are situations where a single skill is all a robot will need to perform, it is not sufficient for general-purpose robots where learning each skill from scratch is impractical. In such cases, there is a great deal of knowledge that can be shared across tasks to speed up learning. In this section, we discuss one particular case study of scalable multi-task learning of vision-based manipulation skills, with a focus on tasks that require pushing or picking and placing objects. Unlike in the previous section, if our goal is to learn many tasks with many objects, a challenge discussed in detail in Section~\ref{subsec:generalization}, it will be most practical to learn from data that can be collected at scale, without human supervision or even a human attending the robot. As a result, it becomes imperative to remove assumptions such as regular resets of the environment or a carefully instrumented environment for measuring reward.

Motivated by these challenges, the visual foresight approach~\citep{finn2017foresight,ebert2018foresight}
leverages large batches of off-policy, autonomously collected experience to train an action-conditioned video prediction model, and then uses this model to plan to accomplish tasks.
The key intuition of this approach is that knowledge learned about physics and dynamics can be shared across tasks and largely decoupled from goal-centric knowledge. These models are trained using streams of robot experience, consisting of the observed camera images and actions taken, without assumptions about reward information. After training, a human provides a goal, by providing an image of the goal or by indicating that an object corresponding to a specified pixel should be moved to a desired position. Then, the robot performs an optimization over action sequences in an effort to minimize the distance between the predicted future and the desired goal.

This algorithm has been used to complete object rearrangement tasks such as grasping an apple and putting it on a plate, re-orienting a stapler, and pushing other objects into configurations~\citep{finn2017foresight,ebert2018foresight}. Further, it has been used for visual reaching tasks~\citep{byravan2018se3posenets}, object pushing and trajectory following tasks~\citep{yen2019experience}, for satisfying relative object positioning tasks~\citep{few_shot_learning_xie18a}, and
for cloth manipulation tasks such as folding shorts, covering an object with a towel, and re-arranging a sleeve of a shirt~\citep{ebert2018foresight}.
Importantly, each collection of tasks can be performed using a single learned model and planning approach, rather than having to re-train a policy for each individual task or object. This generalization precisely results from the algorithms ability to leverage broad, autonomously-collected datasets with hundreds of objects, and the ability to train reusable, task-agnostic models from this data.

Despite these successes, there are a number of limitations and challenges that we highlight here. First, although the data collection process does not require human involvement, it uses a specialized set-up with the robot in front of a bin with tilted edges that ensure that objects \old{to}\new{do} not fall out, along with an action space that is constrained within the bin. This allows continuous, unattended data collection, discussed further in Section~\ref{subsec:robot_at_scale}. Outside of laboratory settings, however, collecting data in unconstrained, open-world environments introduces a number of important challenges, which we discuss in Section~\ref{subsec:robot_persistence}. Second, inaccuracies in the model and reward function can be exploited by the planner, leading to inconsistencies in performance. We discuss these challenges in Sections~\ref{sec:model_based} and~\ref{sec:rewards}. Finally, finding plans for complex tasks pose a challenging optimization problem for the planner, which can be addressed to some degree using demonstrations (for details, see Section~\ref{subsec:handling_exploration}).
This has enabled the models to be used for tool use tasks such as sweeping trash into a dustpan, wiping objects off a plate with a sponge, and hooking out-of-reach objects with a hook~\citep{xie2019improvisation}.

\subsection{Learning to grasp with deep RL}
\label{sec:grasping}

\begin{figure}[th]
\begin{center}
\includegraphics[width=\columnwidth]{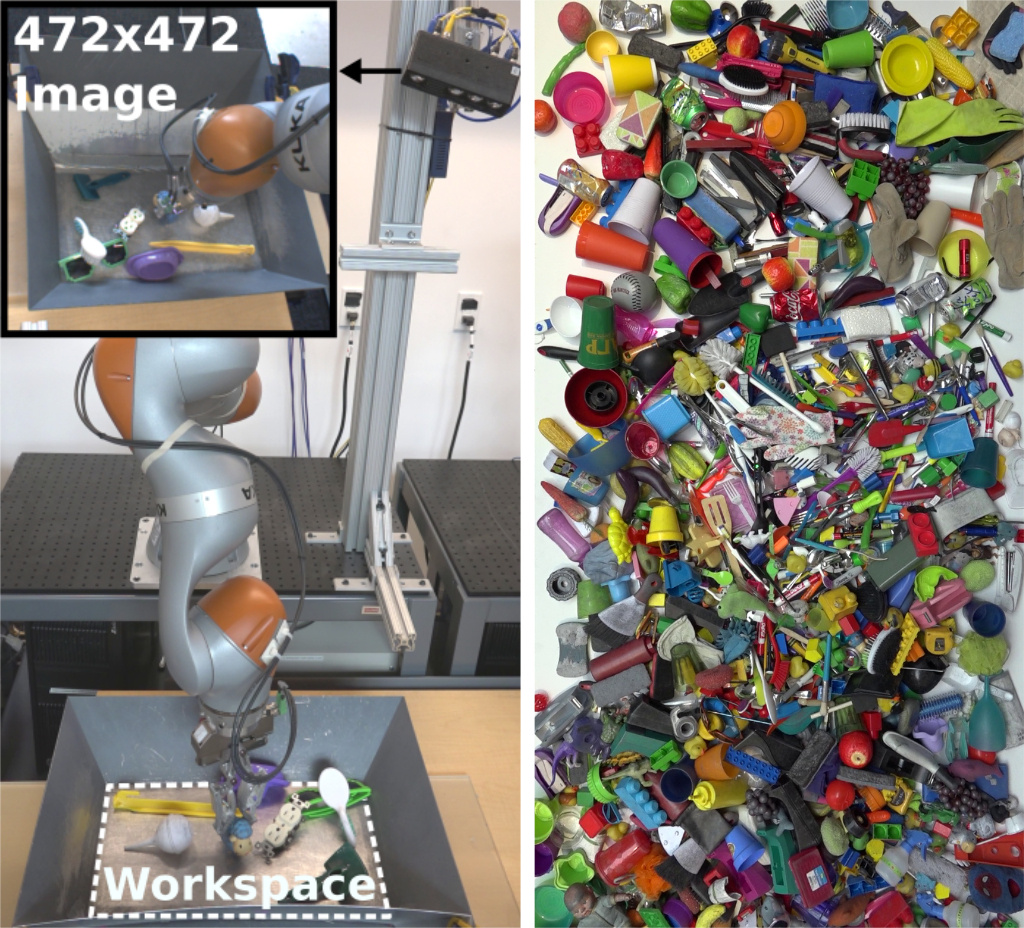}
\caption{Close-up of our robot grasping setup in our setup (left) and about 1000 visually and physically diverse training objects (right). Each robot consists of a KUKA LBR
IIWA arm with a two-finger gripper and an over-the-shoulder RGB camera.
}
\label{fig:grasping}
\end{center}
\end{figure}

Learning to grasp remains one of the most significant open problems in robotics, requiring complex interaction with previously unseen objects, closed loop vision-based control to react to unforeseen dynamics or situations\old{, as well as dexterous manipulation}\new{, and in some cases pre-manipulation to isolate the object to be grasped}. Indeed, most object interaction behaviors require grasping the object as the first step. Prior work typically tackles grasping as the problem of identifying suitable grasp locations~\citep{zeng2018, juxi18, dexnet30_2017, platt_gpd_17}, rather than as an explicit control problem. The motivation for this problem definition is to allow the visual problem to be completely separated from the control problem, which becomes an open loop control problem. This separation significantly simplifies the problem. The drawback is that this approach cannot adapt to dynamic environments or refine its strategy while executing the grasp. Can deep RL provide us with a mechanism to learn to grasp directly from experience, and as a dynamical and interactive process?

\begin{figure*}
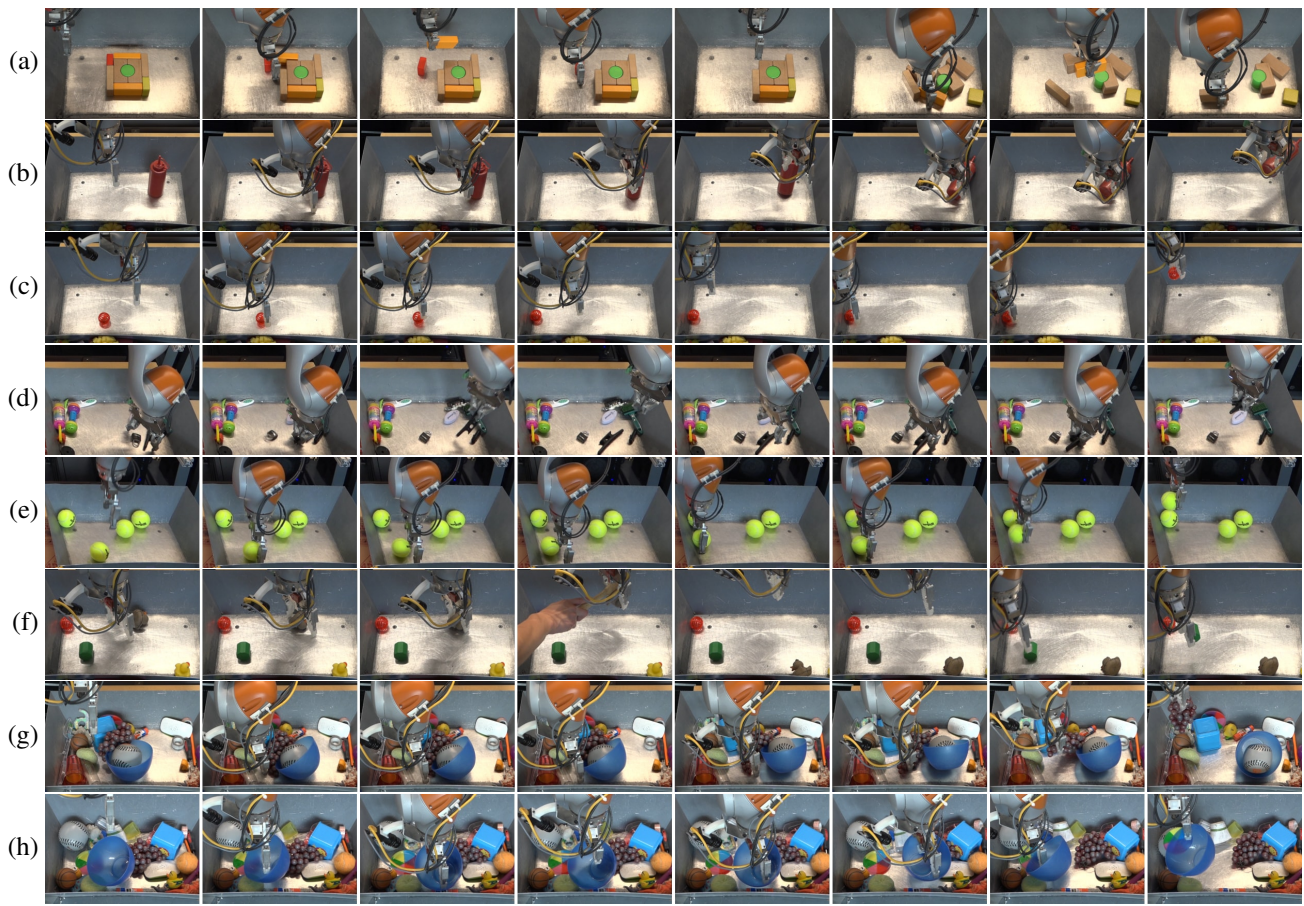

    \vspace{-0.05in}
    \centering
    \begin{imgrows}
        \imgrow \img{images/rows/image1}
        \imgrow \img{images/rows/image2}
        \imgrow \img{images/rows/image3}
        \imgrow \img{images/rows/image4}
        \imgrow \img{images/rows/image5}
        \imgrow \img{images/rows/image6}
        \imgrow \img{images/rows/image7}
        \imgrow \img{images/rows/image8}
    \end{imgrows}
    \caption{Eight grasps from the QT-Opt policy, illustrating some of the strategies discovered by our method: pregrasp manipulation (a, b), grasp readjustment (c, d), grasping dynamic objects and recovery from perturbations (e, f), and grasping in clutter (g, h).}
    \label{fig:qt_opt_qualitative}
    \vspace{-0.2in}
\end{figure*}

A number of works have studied closed loop grasping~\citep{yu2018icra,platt17,hausman17,hand_eye_coordination}. In contrast to these methods, which frame closed-loop grasping as a servoing problem, QT-Opt\new{~\cite{qt_opt}} uses a general-purpose RL algorithm to solve the grasping task, which enables \old{long-horizon reasoning} \new{multi-step reasoning, in other words, the policy can be optimized across the entire trajectory}. In practice, this enables this method to autonomously acquire complex grasping strategies, some of which we illustrate in Figure.~\ref{fig:qt_opt_qualitative}. This method is also entirely self-supervised, using only grasp outcome labels that are obtained automatically by the robot. Several works have proposed self-supervised grasping systems~\citep{pinto16,hand_eye_coordination}, but to the best of the author's knowledge, this method is the first to incorporate \old{long-horizon reasoning} \new{a multi-step optimization} via RL into a generalizable vision-based system trained on self-supervised real-world data.

Related to this work,~\cite{zeng2018} recently proposed a Q-learning framework for combining grasping and pushing. \new{QT-Opt} utilizes a much more \old{generic}\new{flexible} action space, directly commanding gripper motion \new{in all degrees of freedom} in 3 dimensions, and exhibits substantially better performance and generalization. Finally, in contrast to many current grasping systems that utilize depth sensing~\citep{dexnet30_2017, morrison18} or wrist-mounted cameras~\citep{platt17, morrison18}, QT-Opt operates on raw monocular RGB observations from an over-the-shoulder camera \new{that doesn't need to be calibrated}. \old{and the} \new{The} performance of \old{our method}\new{QT-Opt} indicates that effective learning can achieve excellent grasp success rates even with this rudimentary \old{sensor}\new{sensing set-up}.

In this work, we focus on evaluating the success rate of the policy in grasping never seen during training objects in a bin using a top-down grasping (4 degrees of freedom). This task definition simplifies some robot safety challenges, which are discussed more in Section~\ref{subsec:safety_and_reliability_during_training}. However, this problem still retains the challenging aspects that have been hard to deal with: unknown object dynamics, geometry, vision based closed-loop control, self-supervised approach as well as hand eye coordination by removing the need to calibrate the entire system (\new{camera and} gripper locations as well as workspace bounds are not given to the policy).

For this specific task, QT-Opt\old{, a Q-learning based approach} can reach 86\% grasp success when learning completely from data collected from previous experiments which we will refer to as offline data, and can quickly reach 96\% success with an additional online data of 28,000 grasps collected during a joint finetuning training phase. Those results show that RL can be scalable and practical on a real robotic application by either allowing to reuse past collected experiences (offline data), and potentially training purely offline (no additional robot interaction required) or a combination of \new{offline and online} approaches (\new{called} joint finetuning). Leveraging offline data makes deep RL a practical approach for robotics as it allows to scale the training dataset to a large enough size to allow generalization to happen, with a small robotic fleet of 7 robots and over a period of a few months, or by leveraging simulation, to generalize with a collection effort of just a few days~\cite{james2019sim,rao2020rl} \new{(see Section~\ref{subsec:simulation} for more examples of sim-to-real techniques)}.

Because the policy is learned by optimizing the reward across the entire trajectory (optimizing for long term reward using Bellman backup),  and is constantly re-planning its next move with vision as an input, the policy can learn complex behaviors in a self-supervised manner that would have been hard to program, such as singulation, pregrasp manipulation, dealing with a cluttered scene, learning retrial behaviors as well as handling environment disturbance and dynamic objects (Figure ~\ref{fig:qt_opt_qualitative}). Retrial behaviors can be learned because the policy can quickly react to the visual input, at every step, which may show in one step that the object dropped after the gripper lifted it from the bin, and thus deciding to re-attempt a grasp in the new location the object fell to.

Section~\ref{subsec:sample_efficiency} describes some of the design principles we used to get good data efficiency. 
Section~\ref{subsec:generalization} discusses strategies that allowed us to generalize properly to unseen objects. Section~\ref{subsec:robot_at_scale} describes ways we managed to scale to 7 robots with one human operator as well as enable 24h/7 day operations. Section~\ref{subsec:handling_exploration} discusses how we side-stepped exploration challenges by leveraging scripted policies.

The lessons from this work have been that (1) a lot of varied data was required to learn generalizable grasping, which means that we need unattended data collection and a scalable RL pipeline; (2) the need for large and varied data means that we need to leverage all of the previously collected data so far (offline data) and need a framework that makes this easy is crucial; (3) to achieve maximal performance, combining offline data with a small amount of online data allows us to go from 86\% to 96\% grasp success.

\subsection{Learning Legged Locomotion}
\label{subsec:locomotion}

\begin{figure}[th]
\begin{center}
\includegraphics[width=0.5\columnwidth]{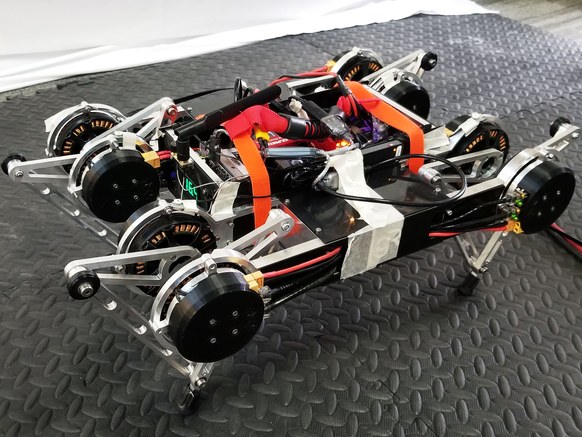}
\caption{The Minitaur robot learns to walk from scratch using deep RL.}
\label{fig:minitaur}
\end{center}
\end{figure}

Although walking and running seems effortless activities for us, designing locomotion controllers for legged robots is a long-standing challenge \citep{Raibert:1986:LRB:6152}. RL holds the promise to automatically design high performance locomotion controllers~\citep{icra04,tedrake2005learning,ha2018automated,Hwangboeaau5872,Lee_2020}. In this case study, we apply deep RL techniques on the Minitaur robot (Figure~\ref{fig:minitaur}), a mechanically simple and low cost quadruped platform \citep{de2017}. We have overcome significant challenges and developed various learning-based approaches, with which agile and stable locomotion gaits emerge automatically. 

Simulation is an important prototyping tool for robotics, which can help to bypass many challenges of learning on real systems, such as data efficiency and safety. In fact, most of the prior work used simulation \citep{BrockmanCPSSTZ16,coumans2017} to evaluate and benchmark the learning algorithms \citep{hamalainen2015online,parkour2017,yu2018learning,peng2018deepmimic}. Using general-purpose RL algorithms and a simple reward for walking fast and efficiently, we can train the quadruped robot to walk in simulation within 2-3 hours. However, a policy learned in simulation usually does not work well on the real robot. This performance gap is known as the \emph{reality gap}. Our research has identified the key causes of this gap and developed various solutions. Please refer to Section~\ref{subsec:simulation} for more details. \old{After narrowing the reality gap} \new{With these sim-to-real transfer techniques}, we can successfully deploy the controllers learned in simulation on the robots with zero or only a handful of real-world experiments \citep{tan2018,yu2019learning}. Without much prior knowledge and manual tuning, the learning algorithm automatically finds policies that are more agile and energy efficient than the controllers developed with the traditional approaches.

Given the initial policies learned in simulation, it is important that the robots can continue their learning process in the real-world in a life-long fashion to adapt their policies to the changing dynamics and operation conditions. There are three main challenges for real-world learning of locomotion skills. The first is sample efficiency. Deep RL often needs tens of millions of data samples to learn meaningful locomotion gaits, which can take months of data collection on the robot. This is further exacerbated by the need for extensive hyperparameter tuning. We have developed novel solutions that have significantly reduced the sample complexity (Section \ref{subsec:sample_efficiency}) and the need for hyperparameter tuning (Section \ref{subsec:stable_learning}). 

Robot safety is another bottleneck for real-world training. During the exploration stage of learning, the robot often tries noisy actuation patterns that cause jerky motions and severe wear-and-tear of the motors. In addition, because the robot has yet to master balancing skills, the repeated falling quickly damages the hardware. We discuss in Section \ref{subsec:safety_and_reliability_during_training} several techniques that we employ to mitigate the safety concerns for learning locomotion with real robots.

The last challenge is \emph{asynchronous control}. On a physical robot, sensor measurements, neural network inference and action execution usually happen simultaneously and asynchronously. The observation that the agent receives may not be the latest owing to computation and communication delays. However, this asynchrony breaks the fundamental assumption of the markovian decision process (MDP). Consequently, the performance of many deep RL algorithms drop dramatically in the presence of asynchronous control. In locomotion tasks, asynchronous control is essential to achieve high control frequency. In other words, to learn to walk, the robot has to think and act at the same time. We discuss our solutions of this challenge in Section \ref{sec:asynchronous}, for both model-free and model-based learning algorithms.

With the progress to overcome these challenges, we have developed an efficient and autonomous on-robot training system \citep{haarnoja2018}, in which the robot can learn walking and turning, from scratch in the real world, with only five minutes of data \citep{yang2019data} and little human supervision.

\section{Outstanding Challenges in Deep RL and Strategies to Mitigate Them}
\label{sec:tameable_challenges}
In the previous section, we showed a few examples of applications of deep RL on robotic tasks that enabled progress over previous approaches in terms of generalization to a large variety of environments, objects or more complex behaviors. Those applications required to solve or at least mitigate a few challenges specific to applying deep RL on real robots that have been identified over the years. In this section, we describe those challenges and provide, whenever available, our current best mitigation strategies that enabled us to apply deep RL to the applications we discussed in Section~\ref{sec:case_studies}.   

\subsection{Reliable and Stable Learning}
\label{subsec:stable_learning}

Deep RL algorithms are notoriously difficult to use in practice~\citep{rlblogpost}. The performance of commonly used RL methods depends on careful settings of the hyperparameters, and often varies substantially between runs (i.e., for different ``random seeds'' in simulation). Off-policy algorithms, which are particularly desirable in robotics owing to their improved sample efficiency, can suffer even more from these issues than on-policy policy gradient methods. We can broadly classify the challenges of reliable and stable learning into two groups: (1) reducing sensitivity to hyperparameters, and (2) reducing issues owing to local optima and delayed rewards.

One approach to reducing the burden of tuning hyperparameters is to use automated hyperparameter tuning methods~\citep{chiang2019learning}. However, such methods typically require running RL algorithms many times, which is impractical outside of simulated domains. A potentially promising alternative available for off-policy RL methods is to run multiple learning processes with different hyperparameters on the same off-policy data buffer, effectively using one run's worth of data for multiple independent learning processes. Recent work has explored this idea in simple simulated domains~\citep{khadka2019collaborative}, though it remains to be seen if such an approach can be scaled up to real-world robotic learning settings. Another approach is to develop algorithms that automatically tune their own hyperparameters, as in the case of SAC with automated temperature tuning, which has been demonstrated to greatly reduce the need for hyperparameter tuning across domains, thus enabling much easier deployment on real-world robotic systems~\citep{haarnoja2018}. Lastly, we can aim to develop methods that are, through their design, more robust to hyperparameter settings. This option, although the most desirable, is also the toughest, because it likely requires an in-depth understanding for the real reasons behind the sensitivity of current RL algorithms, which has so far proven elusive.

The second challenge to reliable and stable learning is local optima and delayed rewards. In contrast to supervised learning problems, which put a convex loss function on top of a nonlinear neural network function approximator, the RL objective itself can present a challenging optimization landscape independently of the policy or value function parameterization, which means that the usual benefits of over-parameterized networks do not fully resolve issues relating to local optima. This is indeed part of the reason why different runs of the same algorithm can produce drastically different solutions, and it presents a major challenge for real-world deployment, where even a single run can be exceptionally time-consuming. Some methods might provide better resilience to local optima by preferring stochastic policies that can explore multiple strategies simultaneously~\citep{ziebart2008maximum,toussaint2009robot,rawlik2013stochastic,fox2015taming,haarnoja2017reinforcement, haarnoja2018soft}.
More sophisticated exploration strategies might further alleviate these issues~\citep{fu2017ex2, pathak2017curiosity},
and parameter-space exploration strategies might offer a particularly promising approach to combating this issue~\citep{burda2018exploration}. Indeed, we have observed in some of our own experiments that, \old{when sample complexity is not required}\new{when collecting large amount of on-policy data is not an issue}, direct parameter search methods such as augmented random search~\citep{mania2018simple} can often be substantially easier to deploy than more classic RL methods, likely to their ability to avoid local optima by exploring directly in the parameter space. It may therefore prove fruitful to investigate methods that combine entropy maximization and parameter space exploration as a way to avoid the local optima and delayed reward issues that make real-world deployment challenging.

\subsection{Sample Efficiency}
\label{subsec:sample_efficiency}
Many popular RL algorithms require millions of stochastic gradient descent (SGD) steps to train policies that can accomplish complex tasks~\citep{schulman2017proximal,mnih2013playing}. This often means that millions of interaction with the real world will be required for robotic tasks, which is quite prohibitive in practice. Without any improvement in sample efficiency to those algorithms, the number of training steps will only increase as the model size increases to tackle more and more complex robotic tasks.

We have found that some classes of RL algorithms are much more sample efficient than others. RL algorithms can be categorized into model-based versus model-free methods. Among the model-free methods, they are often categorized into on-policy and off-policy methods. Generally speaking, among model-free techniques, off-policy methods are about an order of magnitude more data efficient than on-policy methods. Model-based methods could be another order of magnitude more data efficient than their model-free counterparts. In the following sections we discuss our experiences with these methods.

\subsubsection{Off-Policy Algorithms}
On-policy algorithms such as policy gradient methods~\citep{peters2006policy,schulman2015} have recently become popular owing to their stability and their ability to learn policies for a wide variety of tasks. Unfortunately, on-policy algorithms have the constraint to only use a sample coming from the latest policy that is being trained. This has the unfortunate consequence that the number of required data samples is equal and often larger to the number of training steps needed to train a model, which in practice, can be millions of steps. Training an on-policy model may thus require several millions and sometimes billions of action executions in the real world which is often prohibitive.

Off-policy methods do not assume that the samples are coming from the current trained policy. In practice, this means the samples can be reused multiple times across back-propagations, potentially hundreds or thousands of times, without any over-fitting in complex visual tasks. In ~\cite{qt_opt}, up to 15 training steps of batch size 32 were done per collect step on real robots during a finetuning phase, which is equivalent to 480 gradient descents per collect step. Recently, SAC~\citep{haarnoja2018}, an off-policy method, was able to learn to walk on a quadruped robot, from scratch, with just 2 hours of real robot data coming from a single robot. Note that further increasing the ratio between the number of training steps and the number of collected samples may decrease the training performance due to overfitting. The optimal ratio is often task dependent, policy dependent or algorithm dependent, which is an important hyper-parameter to tune.

\subsubsection{Model-Based Algorithms}

Model-based algorithms such as~\cite{draeger1995model} choose the optimal action by leveraging a model of the environment. The agent may learn from the experience generated using this model instead of collected in the real environment. Thus the amount of data required for model-based methods is usually much less than their model-free counterparts. For example, we leveraged such a technique to effectively learn to walk, from scratch, with only a few minutes of real robot data~\citep{nagabandi2018learning, yang2019data}. The downside is that these methods require to have access to such a model, which is often challenging to acquire in practice. We cover model-based techniques in more detail in Section~\ref{sec:model_based}.

\subsubsection{Input Remapping for High-Dimensional Observations}
\label{sec:input_remapping}

When learning from high-dimensional observations, e.g. image observations, learning visual representations can occupy substantial amount of training and sample complexity. 
One trick for addressing this challenge is via \emph{input remapping}. In particular, when policies are trained in a laboratory environment, the true underlying state of the system may be observable during training, even when the policy to be learned must use vision. In these settings, one policy or multiple local policies can be efficiently learned without vision using privileged state information, and these policies can be distilled into a final policy that takes raw observations as input and is trained to produce the output of the non-vision policies. \new{This trick has been successful in a number of settings including robotic manipulation from image pixels~\citep{end2end,Pinto-RSS-18}, autonomous driving~\citep{chen2020learning}, and robotic locomotion from a history of proprioceptive sensor measurements~\citep{lee2020learning}. }

\begin{figure}[th]
\begin{center}
\includegraphics[width=1.0\columnwidth]{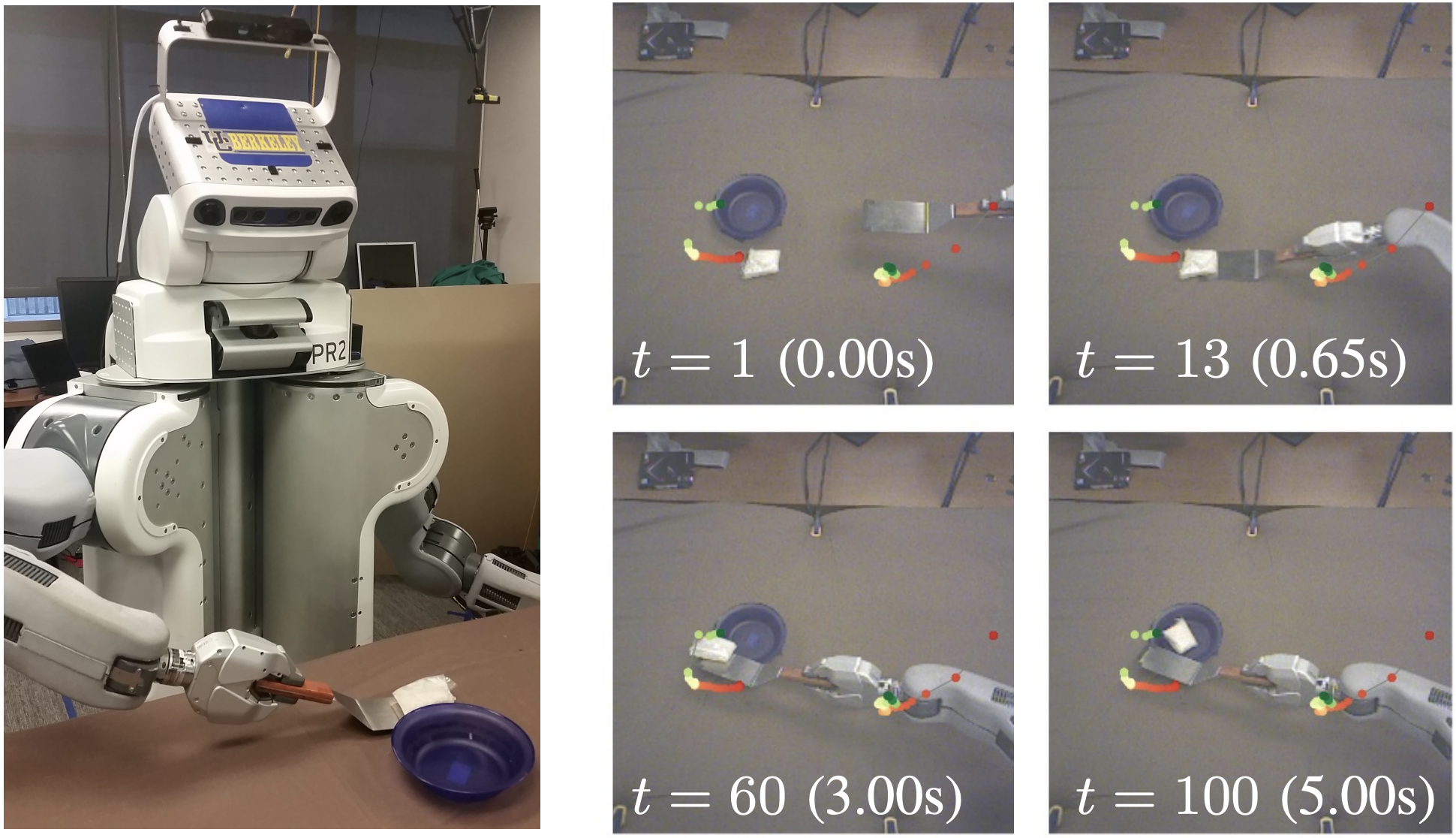}
\caption{PR2 learning to scoop a bag of rice into a bowl with a
spatula (left) using a learned visual state representation (right), using~\citet{dsae}. The feature points visualized on the right images were learned without supervision with an autoencoder.
}
\label{fig:dsae}
\end{center}
\end{figure}

When the true states of objects cannot be measured and the local policies must themselves handle image observations, these observations can be first encoded into a lower-dimensional state space via an autoencoder, such as a spatial autoencoder that summarizes the image with a set of feature points in image-space~\citep{dsae} -- an example of such features are illustrated in Figure~\ref{fig:dsae}.
Unsupervised feature learning methods such as autoencoders~\citep{dsae,ghadirzadeh2017deep}, \new{contrastive losses~\citep{sermanet2018time}, and correspondence learning~\citep{florence2018dense,florence2019self},} provide a reasonable solution in cases where the inductive biases of the unsupervised algorithm effectively match the needs of the state representation. 

\subsubsection{Offline Training}

Image classifiers used by companies, such Facebook or Google, are trained on tens of million of labeled images~\citep{kuznetsova2018open}, or pre-trained on billions of images~\citep{mahajan2018exploring,xie2019self}, to reach the level of quality required by certain products. Natural language processing (NLP) systems for machine translation, or speech recognition systems such as BERT~\citep{devlin2018bert}, also require billions of samples to generalize and have descent performance for real applications. In a way, supervised learning systems are also inefficient, but in many applications, the gains in generalization and performance that deep learning provides compensates for the cost of collecting such large amounts of data. Similarly, a general purpose robot may also require a large volume of data to train on, unless significant improvements have been made in our learning algorithms. Offline training enable us to use all the data collected so far to train our policies, and thus, potentially scale to billions of real world samples.

Off-policy methods can leverage all the data collected in the past, across many experiments. In most RL benchmarks, off-policy methods are still collecting new data as the training happens. However, off-policy methods can also be trained without collecting any new data during the training phase; similar to supervised learning problems. We call this \emph{offline training}, whereas other work may call it batch RL. In Section~\ref{sec:grasping}, we have shown that this mode of training allowed us to generalize grasping policies to unseen objects with just 500,000 trials. If we compare this dataset to the ImageNet dataset, which has about 1 million images, we can see that the amount of data to learn this complex robotic task from vision sensor, using RL, is in the same order of magnitude as learning to classify 1,000 types of objects. In both cases, the learned models have shown the ability to generalize to a wide variety of unseen object instances. There are challenges to stabilize offline training. The offline training can become unstable if the state-action distribution from the latest policy differs too much from the one that was used to collect the training data. Recent work just started to identify and address to some extent this specific problem~\citep{bear,striving_simplicity,fujimoto2018off}.

An important technique to bypass the sample efficiency problem is to use simulators, which can generate realistic experience much faster than real time. Combining with sim-to-real transfer techniques, simulators allow us to learn policies that can be deployed in the real world with a minimal amount of real world interaction. In the next section, we discuss the use of simulation. 

\subsection{Use of Simulation}
\label{subsec:simulation}

Simulation is becoming increasingly accurate over the years, which makes it a good proxy to real robots. One bottleneck of robotic learning is to collect a large amount of data autonomously and safely. While collecting enough real data on the physical system is slow and expensive, simulation can run orders of magnitude faster than real-time, and can start many instances simultaneously. In addition, data can be collected continuously without human intervention. On the real robot, human supervision is always needed for resetting experiments, monitoring hardware status and ensuring safety. In contrast, experiments can be reset automatically, and safety is not a problem in simulation. Thus, prototyping in simulation is faster, cheaper and safer than experimenting on the real robot. These enable fast iteration of developing and tuning learning algorithms. The fast pace of experiments allow us to efficiently shape the reward function, sweep the hyper-parameters, fine-tune the algorithm, and test whether a given task falls within the robot's hardware capability. From our own experience, we have benefited tremendously from prototyping in simulation \citep{tan2018}.

In addition to prototyping, can we directly use the policies trained in simulation on real robots? Unfortunately, deploying these policies can fail catastrophically due to the \emph{reality gap}. Modeling errors cause a mismatch in robot dynamics, and rendered images often do not \new{look} like their real-world counterparts. The reality gap is a major obstacle that prevents the application of learning to robotics. In simulations, the robots can learn to backflip \citep{peng2018deepmimic}, bicycle stunts \citep{tan2014learning}, and even put on clothes \citep{clegg2018learning}. In contrast, \new{it is still very challenging to teach}\old{the} robots \old{fail} to perform basic tasks such as walking in the real world. Bridging the reality gap will allow robotics to fully tap into the power of learning. More importantly, bridging the reality gap is important to push the advancement of machine learning for robotics towards the right direction. In the last few years, the OpenAI Gym benchmark \citep{BrockmanCPSSTZ16} is the key driving force behind the development of deep RL and its application to robotics. However, these simulation benchmarks are considerably easier than their real world equivalent. It does not take into consideration \new{the} detailed dynamics, partial observability, latency, and safety \new{aspects of robotics}. Thus, the scores which researchers optimize their algorithms for can be misleading: the learning algorithms that perform well in the Gym environments may not work well on real robots. If we can bridge this reality gap, we would have a far better simulation benchmark for robotics, which can focus the research effort to the most pressing challenges in robot learning, such as non-Markovian assumption (asynchronous control), partial observability and safe exploration and actuation. In the following section, we outline a few methods that have been employed successfully for sim-to-real transfer.

\new{
\subsubsection{Addressing Partial Observations}
In simulation, we can access the ground-truth state of the robot, which can significantly simplify the learning of tasks. In contrast, in the real-world, we are restricted to \emph{partial} observations that are usually noisy and delayed, due to the limitation of onboard sensors. For example, it is difficult to precisely measure the root translation of a legged robot. To eliminate this difference, we can remove the inaccessible states during training \citep{tan2018}, apply state estimation, add more sensors (e.g. Motion Capture) \citep{haarnoja2018} or learn to infer the missing information (e.g. reward) \citep{Yang2019NoRMLNM}. On the other hand, if used properly, the ground-truth states in simulation can significantly speed up learning. ``Learning by cheating'' \citep{chen2020learning} first leveraged the ground-truth states to learn a privileged agent, and in the second stage, imitated this agent to remove the reliance on the privileged information.}

\subsubsection{Better Simulation}
The reality gap is caused by the discrepancy between the simulation and the real-world physics. This error has many sources, including incorrect physical parameters, un-modeled dynamics, and stochastic real environment. However, there is no general consensus about which of these sources plays a more important role. After a large number of experiments with legged robots, both in simulation and on real robots, we found that the actuator dynamics and the lack of latency modeling are the main causes of the model error. Developing accurate models for the actuator and latency significantly narrow the reality gap \citep{tan2018}. We successfully deployed agile locomotion gaits that are learned in simulation to the real robot without the need for any data collected on the robot.

\subsubsection{Domain Randomization}
The idea behind domain randomization is to randomly sample different simulation parameters while training the RL policy. This can include various dynamics parameters~\citep{Sim2Real2018,tan2018} of the robot and the environment, as well as visual and rendering parameters such as textures and lighting~\citep{sadeghi2016cad2rl,dr_for_sim2real}. Similar to data augmentation methods in supervised learning, policies trained under such diverse conditions tend to be more robust to such variations, and can thus perform better in the real-world.

\subsubsection{Domain Adaptation}

The success of adversarial training methods such as generative adversarial networks (GAN) \citep{NIPS2014_5423} have resulted in their application to several other problems, including sim-to-real transfer. Adapter networks have been trained that convert simulated images to look like their real-world counterparts,
which can then be used to train policies in simulation~\citep{JamesDJ17, shrivastava2017learning, bousmalis2017unsupervised,bousmalis2018using,rao2020rl}.
An alternative approach is that of~\cite{james2019sim} which trains an adapter network to convert real-world images to canonical simulation images, allowing a policy trained only in simulation to be applied in the real-world.
Training of the real-to-sim adapter was achieved by using domain-randomized simulation images as a proxy for real-world images, removing the need for real-world data altogether. The resulting policy achieved 70\% grasp success in the real-world with the QT-Opt algorithm, with no real-world data, and reaches a success rate of 91\% after fine-tuning on just 5,000 real-world grasps: a result which previously took over 500,000 grasps to achieve.

\subsection{Side-Stepping Exploration Challenges}
\label{subsec:handling_exploration}

In RL, ``exploration'' refers most generally to the problem of choosing a policy that allows an agent to discover high-reward regions of the state space. Such a policy may not itself have very high average reward -- typically, good exploration strategies are risk-seeking~\citep{bellemare2016unifying},
highly stochastic~\citep{ziebart2008maximum,toussaint2009robot,rawlik2013stochastic,fox2015taming,osband2016deep,haarnoja2017reinforcement},
and prioritize novelty over exploitation~\citep{bellemare2016unifying,fu2017ex2,pathak2017curiosity}.

In practice, effective exploration is particularly challenging in tasks with \emph{sparse reward}. In the most extreme version of this problem, the agent must essentially find a (high reward) needle in a (zero reward) haystack. Unfortunately, the most natural formulation of many practical robotics tasks has this property. For many tasks, it is most natural to formulate them as \emph{binary reward} tasks~\citep{rlblogpost}: a grasping robot can either succeed or fail at grasping an object, a pouring robot can pour water into a glass or not, and a mobile robot can reach the destination or not. One can reasonably regard these as the most basic task specification, with any more informative reward (e.g., distance to the goal) as additional engineer-provided shaping.

For this reason, a number of prior works have focused on studying exploration for sparse-reward robotic tasks~\citep{andrychowicz2017hindsight,schoettler2019deep}.
Numerous methods for improving exploration have been proposed in the literature~\citep{ziebart2008maximum,toussaint2009robot,rawlik2013stochastic,fox2015taming,osband2016deep,pathak2017curiosity,haarnoja2017reinforcement},
and many of these can be applied directly to real-world robotic RL. However, for certain real-world robotic tasks, this problem can often be side-stepped using a combination of relatively simple manual engineering and demonstration data, and this provides a very powerful mechanism for avoiding a major challenge and instead focusing on other issues, such as efficiency and generalization. The use of demonstrations to mitigate exploration challenges has a long history in robotics~\citep{ins-minds-02,ps-rlmsp-08,dnkp-lsmt-13,mkgp-lsmpd-14},
and has been used in a number of recent works~\citep{jain2019learning,nair2018overcoming}.
There are various ways to incorporate the demonstrations into the learning process, which are discussed in the following section.

\subsubsection{Initialization} A simple way to incorporate demonstrations to mitigate the exploration challenge is to pre-train a policy network with demonstrations via \old{supervised}\new{imitation} learning (also called behavioral cloning) \citep{bojarski2016end}. This approach has been used in a variety of prior robotic learning works~\citep{ins-minds-02,ps-rlmsp-08,dnkp-lsmt-13,mkgp-lsmpd-14}.

Although this approach is simple and often effective, it suffers from two major challenges. First, \old{behavioral cloning}\new{imitation learning} lacks effective guarantees on performance both in theory and in practice~\citep{ross2011reduction},
and the resulting policies can suffer from ``compounding errors,'' where a small mistake throws the policy into an unexpected state, where it makes a bigger mistake. Second, the learned initialization can be easily forgotten by the RL. As it is common practice to begin RL with a high random exploration factor, RL can quickly decimate the pre-trained policy, and end up in a state that is no better than random initialization. Note that some algorithms and policy representations are particularly amenable to initialization from demonstrations. For example, dynamic movement primitives (DMPs) can be initialized from demonstrations in a way that does not suffer from compounding errors~\citep{schaal2006dynamic},
whereas guided policy search can be initialized from demonstration by pre-training the local policies, which in practice tends to be a lot more stable than demonstration pre-training for standard policy gradient or actor-critic methods~\citep{levine1501learning}.

\subsubsection{Data Aggregation} Another technique for incorporating demonstrations in off-policy model-free RL is to add demonstration data to the data buffer for the off-policy algorithm. This method is often used with Q-learning or actor-critic style algorithms~\citep{vevcerik2017leveraging,gao2018reinforcement}.
This can in principle mitigate the exploration challenge, because the algorithm is exposed to high-reward behavior, but tends to be problematic in practice, because commonly used approximate dynamic programming methods (i.e., value function estimation) need to see both good \emph{and} bad experience to learn which actions are desirable. Therefore, when the demonstrations are much better than the agent's own experience, the value function will typically learn that the demonstrated states are better, but might fail to learn which actions must be taken to reach those states. Therefore, this tends to be much more effective when combined with the next method.

\subsubsection{Joint Training} Instead of simply pre-training the policy with supervised learning, we can train it jointly, adding together the loss from the policy gradient objective with the loss for behavioral cloning~\citep{dqfd,gao2018reinforcement,johannink2019residual}. This simple approach provides a much stronger signal to the learner, generally succeeding in staying close to the demonstrations, but at the cost of biasing policy learning: if the demonstrations are suboptimal, the behavioral cloning loss may prevent the RL algorithm from discovering a better policy.

\subsubsection{Demonstrations in Model-Based RL} 
\begin{figure}[t]
\begin{center}
\includegraphics[width=1.0\columnwidth]{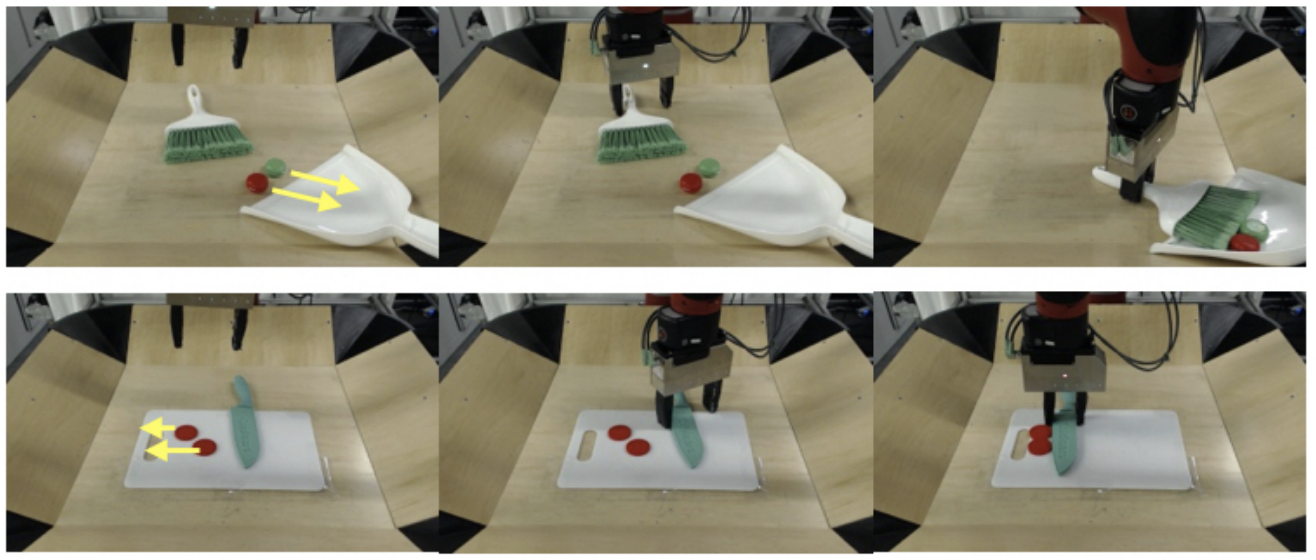}
\caption{Using both unsupervised interaction and teleoperated demonstration data, the robot learns a visual dynamics model and action proposal model that enables it to perform new tasks with novel, previously-unseen tools (using~\citet{xie2019improvisation}). The task specification is shown on the left and the robot performing the task is shown on the right.
}
\label{fig:tool_use}
\end{center}
\end{figure}

In model-based RL, demonstration data can also be aggregated with the agent's experience to produce better models. However, in contrast to the model-free setting, for model-based RL this approach can be quite effective, because it would enable the learned model to capture correct dynamics in important parts of the state space. When combined with a good planning method, which can also use the demonstrations (e.g., as a proposal distribution), including demonstrations into the model training dataset can enable a robot to perform complex behaviors, such as using tools (Figure~\ref{fig:tool_use}), which would be extremely difficult to discover automatically~\citep{xie2019improvisation}.

\subsubsection{Scripted Policies}

In addition to demonstrations, we can also overcome the exploration challenge with a moderate amount of manual engineering, by designing ``scripted'' policies that can serve as initialization. Scripted policies can be incorporated into the learning process in much the same way as demonstrations, and can provide considerable benefit. In the QT-Opt grasping system (Figure~\ref{fig:grasping}), scripted policies are used to pre-populate the data buffer with a higher proportion of successful grasps than would be obtained with purely random actions. Although aggregating such data from a small number of demonstrations would have limited effectiveness, the advantage of a scripted policy is that it can be used to collect very large datasets. In the final QT-Opt experiment, the scripted policy was used to collect 200,000 grasp attempts, with a success rate around 15-30\% Although this success rate is much lower than the final policy, which succeeds 96\% of the time, it was sufficient to bootstrap an effective vision-based grasping skill.

\new{
Another reason why we pre-populate the replay buffer only with a scripted policy is to help keep a ratio of successful and unsuccessful episodes close to 50\%. This is motivated by techniques trying to re-balance equally each class when training a multi-class classifier as in~\cite{chawla2002smote}.  A poor performing policy doesn't generate good data to train a Q-function since it requires both good and bad attempts to be able to learn a good ranking of what a good or bad action is. At the beginning of the training, the policy is bad because the Q-function being learned hasn't converged yet. Such a policy only generates unsuccessful episodes which can't be used to train a good Q-function. This is why the policy is only used to generate data once it reached a certain amount of performance. Our rule of thumb is to only start using the trained policy for data collection once it has reached 20+\% success.}

Scripted policies can also be used in a ``residual'' RL framework, which serves a similar purpose as joint training with demonstrations. In residual RL~\citep{silver2018residual,tan2018,johannink2019residual},
the reinforcement learner learns a policy that is additively combined with the scripted policy, i.e. $\pi_\text{final}(s) = \pi_\text{scripted}(s) + \pi_\text{learned}(s)$. The motivation is similar: unlike pure initialization, the residual approach always retains the scripted component. However, unlike joint training with demonstrations, residual RL can overcome the bias in the scripted policy by learning to ``undo'' $\pi_\text{scripted}(s)$, and therefore can in principle still converge to the optimal policy.

\subsubsection{Reward Shaping}

\new{
Shaping the reward function can also side-step exploration challenges by providing the RL algorithm with additional guidance for exploration. For example, for a reaching task, one can use the distance of the agent to the goal as a negative reward which will significantly speed up the exploration. We've used this approach in several works for learning manipulation skills, such as door opening and peg insertion, where object location information is available during training~\citep{gu2017deep,end2end}. This approach is very effective for any tasks where the agent has to go to a specific know location, such as in navigation tasks~\cite{prmrl}. However, we've found in practice that such an approach can be difficult to scale to many diverse manipulation tasks. This is due to two factors. First, it can be very difficult to weight the shaping terms properly to avoid any greedy and unintentional sub-optimal behaviors. For example, to open the door, one may want to get close to the handle, but may require to take some distance from it to take a different approach with the gripper if the handle can't be moved with the current orientation. Such behavior would go against the shaping of the reward, and thus the reward shaping may make it impossible to discover such a behavior if its  weight is too high. Second, and perhaps more importantly in real-world environments, such shaped reward functions require knowledge of the precise state of the environment, such as object locations relative to the robot.

This is feasible in simulation but can be very challenging on real robots, where the only input may be an image. Once one wants to tackle multiple manipulation tasks, dealing with those variations may be difficult to program even in simulation, since the state configuration one has to deal with can grow exponentially.}

While we have discussed how the challenge of exploration can be side-stepped by employing demonstrations, scripted policies, and reward shaping, the study of exploration and curiosity in robotic learning still plays an important role. Indeed, we can regard \old{demonstrations and scripted policies}\new{those approaches} as a means to parallelize research on robotic learning: if we aim to study perception, generalization, and complex tasks, we can avoid needing to solve exploration as a prerequisite.

\subsection{Generalization}
\label{subsec:generalization}

Generalization to any new skills, environments or tasks still remains an unsolved problem. Solving this problem is required to allow robots to operate in a wide variety of real-world scenarios. However, there are a few restricted situations where we have seen good generalization. In the next section, we cover two important aspects: (1) good data diversity guaranteeing to cover the space we want to generalize and (2) having a correct train and test evaluation protocol that allows us to optimize our system towards better generalization.

\subsubsection{Data Diversity}

Good data diversity that covers the space of generalization we care about is critical to have good performance with deep learning. Deep RL is no exception. In QT-Opt, we cared about generalization to the objects that were never seen during training. Thus, we made sure that during data collection, the agent would see more than 1,000 different object types. If we had only collected data with a small set of objects, we may not achieve the generalization capability that we need. It is the same analogy that we cannot expect a model trained on CIFAR (with 100 classes) to generalize as well as a model trained on ImageNet (with 1,000 classes). This is also true for robotics. If we want to generalize to any objects, we may need to collect data with thousands of them. If we want the policy to be agnostic to the robot arm geometry, we may need to train with thousands of arm variations, etc.

A lot of recent work has leveraged domain randomization in simulation to obtain good sim-to-real transfer because they cared about generalization to a new environment. There is a tradeoff here as more environment diversity may cause the policies to have lower performance. Often this can be alleviated with larger and better neural network architectures. As an example, a larger and deeper than usual neural network was required in~\cite{qt_opt} for the Q-function to deal with the large variety of objects and to achieve good performance on test objects.

\subsubsection{Proper Evaluation}

To get good generalization, the entire system, including its hyperparameters, has to be tuned to optimize for it. This means that when we define the evaluation protocol, we have to be thoughtful to have two MDPs: one for training, and a separate one for evaluation.

This separation of MDP has to be done based on what we care to generalize against: if we want a policy that can grasp new objects, we should have the training MDP with a different set of objects than the testing MDP, both in simulation and the real setup. If we care about generalizing to new robot dynamics, we should make sure to define our training MDP with different dynamics than our testing MDP.

\subsection{Avoiding Model Exploitation}
\label{sec:model_based}

There have been notable success stories in robotics with model-based RL approaches that learn a model of the dynamics and use that model to choose actions~\citep{deisenroth2013survey,end2end,lenz2015deepmpc,finn2017foresight,nagabandi2018learning,xie2019improvisation,thanard_metrpo,yang2019data}.
Here, we use the term `model-based' to describe algorithms that learn a model of the dynamics from data, not to refer to the setting where a model is known a priori. Empirically, these methods have enjoyed superior sample complexity in comparison to model-free approaches~\citep{deisenroth2013survey,nagabandi2018learning,yang2019data}, have scaled to vision-based tasks~\citep{end2end,dsae,finn2017foresight}, and demonstrated  generalization capabilities to many objects and tasks when the model is trained on large, diverse datasets~\citep{finn2017foresight,yang2019data}. These
generalization capabilities are a natural byproduct of being able to train on off-policy datasets. 

Despite the benefits of model-based RL methods, a primary, well-known challenge faced by such model-based RL approaches is model exploitation, i.e. when the model is imperfect in some parts of the state space, and the optimization over actions finds parts of the state space where the model is erroneously optimistic. This can result in poor action selection. Although this challenge is real, we have found that, in practice, we have multiple tools for mitigating it.

First, we have found that optimization under the model is successful when the data distribution consists of particularly broad distributions over actions and states~\citep{finn2017foresight}. In problem domains where this is not possible, one effective tool is data aggregation, which interleaves the data collection and model learning, similar to DAGGER \citep{ross2011reduction}. Whenever the model is inaccurate and gets exploited, more data in the real world is collected to re-train the model. Another tool is to represent and account for model uncertainty~\citep{DeisenrothRT2011}. Acquiring accurate uncertainty estimates when using neural network models is particularly challenging, though there has been some success on physical robots~\citep{nagabandi2019deep}. If we cannot obtain uncertainty estimates, then we can alternatively model the data distribution that the model was fit, and constrain the optimization to that distribution. We have found this approach to be particularly effective when using models fit locally around a relatively small number of trajectories~\citep{end2end,pigps}. We can achieve a similar effect, but without having to refit models from scratch, by learning to adapt models to local contexts from a few transitions~\citep{nagabandi2018grbal}: this approach allows us to automatically construct local models from short windows of experience. These local models have been demonstrated on a variety of robotic manipulation and locomotion problems.

Even if the learned model is accurate for a single-step prediction, error can accumulate over the a long-horizon plan. 
For example, the predicted and real trajectories can quickly diverge after a contact event, even if the single-step model error is small. We found that using multi-step losses~\citep{finn2017foresight,yang2019data}, shorter horizons (when applicable)~\citep{nagabandi2018learning} and replanning~\citep{finn2017foresight,nagabandi2018learning} are effective strategies for limiting the error accumulation, and recovering from model exploitation.

\subsection{Robot Operation at Scale}
\label{subsec:robot_at_scale}

Recent advances in deep learning have also contributed to faster compute architectures and the availability of ever growing (labeled) data sets~\citep{garofolo1993darpa,deng2009imagenet}. In addition, various open-source efforts, such as those of~\cite{pytorch} and~\cite{tensorflow2015-whitepaper}, have contributed to minimizing the cost of entry. Importantly, progress was enabled also because the time it took to train deep models and iterate on them became shorter and shorter. This holds true for robotic learning as well. The faster training data can be collected and a hypothesis can be tested, the faster progress will be made.

Despite advances in data-efficiency (Section ~\ref{subsec:sample_efficiency}), deep RL still requires a fair amount of data, especially if visual information (images) is part of the observation. 
The majority of robot learning experiments to date were conducted on a single robot closely monitored by a single human operator. This one-to-one relation between robot and operator has been a tedious but effective way to ensure continuous and safe operation. The human can reset the scene, stop the robot in unsafe situations, and simply restart and reset the robot on failures. However, to scale up data collection efforts and increase the throughput of evaluation runs, robots need to run without human supervision. It is impractical to allocate more operators to a setup with multiple robots, or whenever a single robot is meant to run 24h/7, and especially both.
In the following we discuss the particular challenges that arise in those settings, namely (1) designing the experimental setup to maximize throughput, i.e. the number of episodes/trials per hour, (2) facilitating continuous operation of the robots, and (3) dealing with non-stationarity due to environment changes.

\subsubsection{Experiment design}

The experimental setup itself, i.e. how a particular robot is set up to tackle a specific task, is an important and often overlooked aspect of a successful experiment. Oftentimes the setup has been carefully engineered or the task has been chosen such that the robot can reset the scene to facilitate unattended and potentially round-the-clock operation. \new{For example, in~\citep{pinto16,hand_eye_coordination,qt_opt,zeng2018,cabi2019framework,dasari2019robonet}, the workspaces are convex, the objects involved allow for safe interaction, and action-spaces are mostly restricted to top-down combined with either intrinsic compliance of the robot itself and/or a wrist mounted force-torque sensor to detect and stop unsafe motions.}
Ideally, the experimental setup is as unconstrained as possible, but in practice is restricted to create a safe action space for the robot (see Section~\ref{subsubsec:designing_safe_action_spaces}).

\subsubsection{Facilitating continuous operation}

Round-the-clock operation will stress the robot itself as well as the experimental setup. Repeated potentially unintended contact of the robot with objects and environment will wear out any experimental setup eventually and needs to be considered upfront. The challenge for long running experiments is to increase the mean-time-between-failure while ensuring that the data that is being collected is indeed useful for training. The former requires to understand the root cause for each intervention and develop fail-safe redundancies. We discuss this challenge more in Section~\ref{subsec:robot_persistence}. Similarly, to ensure that the collected data is not compromised, adding sanity checks is recommended along with actually using the data early to train and re-train models.
Despite simply acquiring more data faster, running experiments around-the-clock also ensures that robots are exposed to varying amounts of lighting conditions allowing us to train more robust policies. However, spot-checking the collected data is important as we noticed, for example that the ceiling lights automatically turned off for parts of the night resulting in very dark images compromising the data.

\subsubsection{Non-Stationarity due to Environment Changes}
\label{subsubsec:non_stationarity}
A learned policy will fail if environment aspects have significantly changed since training. For example, the lighting conditions may significantly shift at night if windows are present in the room, and evaluations done at night may have very different results if no data collection happened at that time. The underlying dynamics may have shifted significantly since training due to hardware degradation. Hardware degradation, such as change of battery level, wear and tear, and hardware failure, are the major causes of dynamic changes. Traditional learning-based approaches, which have distinctive training and testing phases, assuming stationary distribution between phases, suffer from hardware degradation or environment changes not captured in the collection phase. In extreme cases of locomotion, a learned policy can stop working after merely a few weeks due to significant robot dynamic changes. To address these challenges, learning algorithms need to adjust online \citep{yu2017}, optimize for quick adaptation \citep{finn2017model,Yang2019NoRMLNM}, or learn in a lifelong fashion.

This can also have consequences for evaluation protocols where comparing two learned policies or even the same one at different times. We recently found that the best policy learned in ~\cite{hand_eye_coordination} was sensitive to a hardware degradation of the fingers, which caused a consistent performance drop of 5\% in as little as 800 grasps executed on a single robot. One way to mitigate this is to use proper A/B testing protocols as described in~\cite{a_b_testing}.

\subsection{Asynchronous Control: Thinking and Acting at the Same Time}
\label{sec:asynchronous}

The MDP formulation assumes synchronous execution: the observed state remains unchanged until the action is applied. However, on real robotic systems, the execution is asynchronous. The state of the robot is continuously evolving as the state is measured, transmitted, the action calculated and applied. \emph{Latency} measures the delay from when the observation is measured at the sensor, to when the action is actually executed at the actuator. This delay is usually on the order of milliseconds to seconds, depending on the hardware and the complexity of the policy. The existence of latency means that the next state of the system does not directly depend on the measured state, but instead on the state after a delay of latency after the measurement, which is not observable. Latency violates the most fundamental assumption of MDP~\citep{xiao2020thinking}, and thus can cause failure to some RL algorithms. For example, we tested soft actor-critic (SAC)~\citep{haarnoja2018soft} and QT-Opt~\citep{xiao2020thinking}, two state-of-the-art off-policy algorithms, to learn walking on a simulated quadruped robot or grasping objects with an arm, with different latencies. Although both QT-Opt and SAC can learn efficiently when the latency is zero, they failed when we increase the latency.

Clearly, we need special treatments to combat the non-Markovianness introduced by latency. For model-based methods, the planning component is often computationally expensive, and incurs additional latency. For example, the popular sample-based planner, cross-entropy method (CEM)~\citep{de2005tutorial}, needs to rollout many trajectories and update the underlying distribution of optimal action sequences. Even if CEM is parallelized using the latest GPU, planning alone can still take tens of milliseconds. To accommodate such latency, in~\cite{yang2019data}, we plan the optimal action sequence based on a future state, which is predicted using the learned dynamic model, to compensate for the latency caused by the planning algorithm. For model-free methods, one approach is to add recurrence to the policy network, and in particular, include the previous actions taken by the policy as part of the state definition. The recurrent neural network could learn to extrapolate the observation to when the action is applied, from the memorized previous observations. Another approach along the same line, which avoids the additional cost of training recurrent neural networks, is to augment the observation space with a window of previous observations and actions. In practice, we find that the latter is simpler and equally effective~\citep{haarnoja2018soft,xiao2020thinking}.

\subsection{Setting Goals and Specifying Rewards}
\label{sec:rewards}

A critical component required for any application of RL is the reward function. In simulation or video game environments, the reward function is typically easy to specify, because one has full access to the simulator or game state, and can determine whether the task was successfully completed or access the score of the game.
In the real world, however, assigning a score to quantify how well a task was completed can be a challenging perceptual problem of its own. In most of our case studies, we sidestep this difficulty in one of the following ways. (1) Instrumenting the environment with additional sensors that provide reward information. For example, an inertial measurement unit was used to measure the angle of the door and the handle to learn a door-opening task in ~\cite{pigps}, or a motion capture device was used to measure how fast the quadruped robot walks \citep{haarnoja2018}.
(2) Simple heuristics such as image subtraction or target joint encoder values can be valuable in some cases. For example, \cite{qt_opt} used the gripper encoder values and a comparison of images with and without the grasping in order to determine whether an object was successfully grasped.
(3) Learning a visual prediction model as in \cite{finn2017foresight} avoids the need to define reward functions at training time: instead, the reward is specified at evaluation time based on a goal image or equivalent representation.
However, none of these methods necessarily generalizes to any possible robot task one might wish to solve using RL.

Learning the reward function itself is a promising avenue for addressing this problem. It can be learned explicitly, from demonstrations~\citep{finn2016guided}, from human annotation \citep{cabi2019framework}, \new{from human preferences~\citep{sadigh2017active,christiano2017deep}, or from multiple sources of human feedback~\citep{biyik2020learning}}. 
In these examples, reward function learning is typically done in parallel with the RL process, because new experience data helps train a better reward function approximation. However, large amounts of demonstrations or annotations may be required. The process of learning reward functions from demonstrations, called \emph{inverse RL} is an underspecified problem~\cite{ziebart2008maximum}, making it difficult to scale to image observations, and exploitation of the reward can happen even with in-the-loop reward learning. There are promising techniques to try to address some of these problems, \new{including using meta-learned priors~\citep{few_shot_learning_xie18a} or active queries~\citep{singh2019end}}, but learning rewards with minimal human supervision in the general case remains an unsolved problem.

\subsection{Multi-Task Learning and Meta-Learning}
\label{subsec:multi_task_and_meta}

One promising approach towards enabling robots to learn tasks efficiently is to leverage previous experience from other tasks rather than training for a task completely from scratch. Multi-task learning approaches aim to do exactly this by learning multiple tasks at once, rather than training for a single task. Similarly, meta-learning algorithms train across multiple tasks such that learning a new future task can be done very efficiently. Although these approaches have shown considerable promise in enabling robots to quickly adapt to new object configurations~\citep{duan2017one}, new objects~\citep{finn2017one,james2018task}, and new terrains or environment conditions~\citep{nagabandi2018grbal,yu2019learning}, a number of challenges remain in order to make them practical for learning across many different robotic control tasks in the real world.

The first challenge is to specify the task collection. These algorithms assume a collection of training tasks that are representative of the kinds of tasks that the robot must generalize or adapt to at test time. However, specifying a reward function for a single task already presents a major challenge (Section~\ref{sec:rewards}), let alone for tens or hundreds of tasks. Some prior works have proposed solutions to this problem by deriving goals or skills in an unsupervised manner~\citep{gregor2016variational,jabri2019unsupervised}. However, we have yet to see these approaches show
significant success in real world settings.

Another significant challenge lies in the optimization landscape of multiple tasks. Learning multiple tasks at once can present a challenge even for supervised learning problems due to different tasks being learned at different rates~\citep{gradnorm,schaul2019ray} or the challenges in determining how to resolve conflicting gradient signals between tasks~\citep{sener2018multi}. These optimization challenges can be exacerbated in RL settings, where they are confounded with challenges in trading off exploration and exploitation. These challenges are less severe for similar tasks~\citep{duan2016rl,finn2017model,rakelly2019efficient},
but pose a major challenge for more distinct tasks~\citep{parisotto2015actor,rusu2015policy}.

Finally, as we scale learning algorithms towards many different tasks, all of the existing challenges discussed above remain and can be even more tricky, including the need for resetting the environment towards state that are relevant for the current task~\ref{subsec:robot_persistence}, operating robots at scale~\ref{subsec:robot_at_scale}, and handling non-stationarity~\ref{subsubsec:non_stationarity}.

\subsection{Safe Learning}
\label{subsec:safety_and_reliability_during_training}

Safety is critical when we apply RL on real robots. Although sufficient exploration leads to more efficient learning, directly exploring in the real world is not always safe. Repeated falling, self-collisions, jerky actuation, and collisions with obstacles may damage the robot and its surroundings, which will require costly repairs and manual interventions (Section~\ref{subsec:robot_persistence}).

\subsubsection{Designing Safe Action Spaces}
\label{subsubsec:designing_safe_action_spaces}
One simple way to avoid unsafe behaviors is to restrict the action space such that any action that a learned policy can take is safe. This is usually very restrictive and cannot be applied to all applications. However, there are many cases, particularly in semi-static environments and tasks, such as grasping and manipulation, where this is the right approach. Grasping objects in a bin is a very common task in logistics. In these settings, safety can typically be enforced by restricting the work space.  For example, in ~\cite{hand_eye_coordination} and ~\cite{qt_opt}, all actions are selected through sampling, and unsafe samples are rejected. This allows us to perform safety checks or add constraints to the action space. By using a geometric model of the robot and the world, we can reject actions that are outside the 3D volume above the bin, and reject actions that violate kinematic or geometric constraints. We can also enforce constraints on the velocity of the arm.

Although this allows us to handle safety for parts of the robot and environment that can be modeled, it does not deal with anything that is unmodeled, such as objects in the scene that we might want to grasp or push aside before grasping. We can mitigate this issue by using a force-torque sensor at the end-effector to detect and stop motion when an impact occurs. From the point of view of the RL agent, this action appears to have a truncated effect. This combination of strategies can provide for a workable level of safety in a simple and effective way for tasks that are quasi-static in nature.

\subsubsection{Smooth Actions}
Typically, exploration strategies are realized by adding random noise to the actions. Uncorrelated random noise injected in the action space for exploration can cause jerky motions, which may damage the gearbox and the actuators, and thus is unsafe to execute on the robot. Options for smoothing out jerky actions during exploration include: reward shaping by penalizing jerkiness of the motion, mimicking smooth reference trajectories \citep{peng2018deepmimic}, learning an additive feedback together with a trajectory generator \citep{iscen2018}, sampling temporal coherent noise \citep{haarnoja2018,yang2019data}, or smoothing the action sequence with low-pass filters. All these techniques work well, although additional manual tuning or user-specified data may be required.

In the locomotion case study (Section \ref{subsec:locomotion}), because the learning algorithm can freely explore the policy space, the converged gait may not be periodic, may be jerky or may use too much energy, which can damage the robot and its surroundings. They usually do not resemble the gaits of animals that we are familiar with in nature. Although it is possible to mitigate these problems by shaping the reward function, we find that a better alternative that requires less tuning is to incorporate a periodic and smooth trajectory generator into the learning process. We develop a novel neural network architecture, policies modulated trajectory generator (PMTG) ~\citep{iscen2018}, which can effectively incorporate prior knowledge of locomotion and regularize the learned gait. PMTG subdivides the controller into an open-loop and a feedback component. The open-loop trajectory generator creates smooth and periodic leg motion, whereas the feedback policy, represented by a neural network, can be learned to modulate this trajectory generator, to change walking speed, direction and style. As a result, the PMTG policies are safe to be deployed or directly learned on the real robot.

\subsubsection{Recognizing Unsafe Situations}
It is crucial to recognize that unsafe situations is about to happen, so that a recovering policy can be deployed to keep the robot safe, or to shutdown the robot completely. Heuristic-based approaches can be designed to recognize these unsafe states or actions by checking whether the action will cause collision, or whether the power and the torque exceed the limit. Performing these rule-based safety checks often require careful tuning and a rich set of onboard sensors. Furthermore, we can also employ learning to recognize unsafe situations. These approaches can use ensemble models to estimate uncertainty~\citep{DeisenrothRT2011,eysenbach2018leave} of certain predictions, which can be a good indicator whether any unsafe behavior may happen, \new{or can directly learn the probability of future unsafe behaviors from experience~\citep{gandhi2017learning,srinivasan2020learning}}. Once a precarious situation is recognized, a recovering policy can be deployed to move the robot back to a safe state. The task policy, the recovering policy and the classifier for safety can all be learned simultaneously~\citep{eysenbach2018leave,thananjeyan2020recovery}. For example, in a locomotion task, when the robot is in a balance state, the task policy (walking) is executed and updated. When the robot is about to fall, which is predicted by the learned Q function, the recovering policy (stand up) takes over. The data collected in this mode is used to update the recovering policy. We showed that learning them simultaneously can dramatically reduce the number of falls during training.

\subsubsection{Constraining Learned Policies}
One obvious way to avoid unsafe behaviors is to penalize unsafe actions each time they are taken. However, this can be hard in practice, as careful tuning is needed for the weights of this penalty term. A more effective alternative is to formulate safe RL as a constrained markov decision process (C-MDP) \citep{altman1999constrained}. For example, TRPO ~\citep{schulman2015} ensures a stable learning using a KL divergence constraint. More recently, \citep{achiam2017constrained, bohez2019value} have also applied constraint-based optimization to \old{treat}\new{model} safety as \new{a set of} hard constraints. In our locomotion projects, we formulated a C-MDP that has inequality constraints on the roll and the pitch of the robot base, which constitutes a rough measure of balance. If the state of the robot stays within the constraints throughout the entire training process, the robot is guaranteed to stay upright. This minimizes the chance of falling when the robot is learning to walk. The constrained formulation usually performs better because as long as the constraints are met, no gradients is generated, and thus no interference can happen between the safety constraints and the reward objective. However, too stringent constraints will limit exploration and can lead to slow learning. 

\subsubsection{Robustness to Unseen Observations}
Last but not least, even if the training process is safe, the final learned policy can execute unexpected, and potentially unsafe, actions when encountering unseen observations. To improve the generalization of the policy to unseen situations, we adopted a robust control approach. We use domain randomization, which samples different physical parameters, or add perturbation forces, either randomly or adversarially \citep{pinto2017robust}, to the robot during training, to force it to learn to react under a wide variety of observations. Before deploying the policy on the robots, we also perform extensive evaluations in simulation about the safety and the performance of the controller on untrained scenarios. Occasionally, the robot, which is trained to be robust and passed all the safety checks in simulation, can still misbehave in the real world. In these rare situations, the model-based or heuristic-based safety checks, such as self-collision detection, power/torque limit, acceleration threshold, etc., will trigger and shut down the robot.

\subsection{Robot Persistence}
\label{subsec:robot_persistence}

We use the term \emph{robot persistence} to refer to the capability of the robot to persist in collecting data and training with minimal human intervention. Persistence is crucial for larger-scale robotic learning, because the effectiveness of modern machine learning models (i.e., deep neural networks) is critically dependent on the quantity and diversity of training data, and persistence is required to collect large training sets. We can divide the problem of robot persistence into two main categories: (1) \textit{self persistence} -- the robot must avoid damaging itself during training; (2) \textit{task persistence} -- the robot must act so that it can continue to perform the task. Robot persistence is critical for enabling autonomous data collection safely and at scale.

\subsubsection{Self Persistence}
We define \emph{self-persistence} as the ability for the robot to keep its full range of motion while performing a task. If the robot were to collide with itself, or the environment, and end up damaging itself, it may end up losing certain abilities, requiring human intervention. In Section~\ref{subsec:safety_and_reliability_during_training}, we provide a few strategies to improve self-persistence.

\subsubsection{Task Persistence}

Task persistence is the capability of the robotic setup to accomplish a range of tasks, repeatedly, in the case of grasping, hundreds of thousands of times to learn the task. Being able to retry a task is tightly coupled with the environment itself and is to this day, still an unsolved problem for a large range of tasks.

Challenges can occur where the robot work-space is limited, and thus objects required to accomplish a task may accidentally be thrown out of reach. In this case, we need to find exploration strategies that avoid ending up in such states, in a very limited data regime, to avoid human interventions that are needed every time such an unrecoverable state is encountered. In high dimensional states, such as images, this becomes a challenging problem as even defining those states becomes a challenge on its own: how do we know from an image that an object has fallen off the bin?

Another class of challenges that we also put in this category is what is \old{others may}\new{often called} ``environment reset''. In many cases, once the task is accomplished, changes in the environment may need to happen before another trial can be done. This is easy to do in simulation: just reset the state of the environment. In the real world, this can often be much harder to accomplish, as resetting the environment is a sequence of robotic tasks, which may be as hard or harder than the task we are trying to learn itself. An example is learning to screw the cap of a bottle again, we may have to unscrew it to be able to try to screw it again. Pouring or assembly tasks are also examples where resetting the environment may be as challenging or may require many steps to accomplish. Automating the whole process of environment reset is required if we want the robot to persist to learn the task. It becomes a challenge of identifying the right set of sub-tasks whose reset action we already learned how to do with a robot.

On occasion, some tasks are physically irreversible, such as welding two pieces of metal, cutting food with a knife, cutting paper with scissors, or writing with a marker. In those cases, other robots may have to bring new objects to the robot trying to learn those tasks, which may be much harder than trying to accomplish the task itself.

Solving task persistence remains mostly an open problem. 
Although guided policy search methods that can handle random initial states have been developed~\citep{montgomery2016guided, montgomery2017reset}, they still rely on clustering the initial states into a discrete set of ``similar'' states, which may be impractical in some cases, such as the diverse grasping task discussed in Section~\ref{sec:grasping} and the diverse pushing task in Section~\ref{subsec:foresight}, where the ``state'' includes the positions and identities of all objects in the scene.
Previous work such as~\cite{pinto16,finn2017foresight} limited the task and action space to be within a bin, which helped keep objects in it by having raised side walls as well as tackled tasks that required a simple reset: just open the gripper above the bin and bring it back to a home position which can easily be scripted. Because task persistence was resolved to some extent, some of those work managed to collect millions of trials~\citep{hand_eye_coordination,qt_opt}. Unfortunately, many tasks do not have these nice properties. For example,~\cite{pilqr} leveraged a human to perform the reset by bringing the puck back to a position where the hockey stick could hit it again.~\cite{haarnoja2018} had to bring the legged robot back to its initial starting position every time the robot reached the end of the limited 5m workspace. In both cases, task persistence was not achieved and humans were performing the reset procedure. This makes data collection hard to scale because (1) it was very time consuming for a human and (2) in both cases, they stopped because they started to feel back pain while performing the environment reset. As such, only a few hours of data, and less than 1,000 trials were performed.

More recently, work such as that of \cite{eysenbach2018leave} tried to tackle this issue of task persistence by integrating environment reset as part of the learning procedure, in a task-agnostic way. However this work only explored tasks which have a unique starting point, that can be reached from most states. This strategy is not always possible such as in self-driving cars, where going backward to come back to the starting point is generally not safe.

\section{Discussion and Conclusions}
\label{sec:discussion}

In this article, we discussed how deep RL algorithms can be approached in a robotics context. We provided a brief review of recent work on this topic, a more in-depth discussion focusing on a set of case studies, and a discussion of the major challenges in deep RL as it pertains to real-world robotic control. Our aim was to present the reader with a high-level summary of the capabilities of current deep RL methods in the robotics domain, discuss which issues make deployment of deep RL methods difficult, and provide a perspective on how some of those difficulties can be mitigated or avoided.

Although deep RL is often regarded as being too inefficient for real-world learning scenarios, described in Section~\ref{subsec:sample_efficiency}, we discuss how in fact deep RL methods have been applied successfully on tasks ranging from quadrupedal walking, to grasping novel objects, to learning varied and complex manipulation skills. These case studies illustrate that deep RL can in fact be used to learn directly in the real wold, can learn to utilize raw sensory modalities such as camera images, and can learn tasks that present a substantial physical challenge, such as walking and dexterous manipulation. Most importantly, these case studies illustrate that policies trained with deep RL can generalize effectively, such as in the case of the robotic grasping experiments discussed in Section~\ref{sec:grasping}.

However, utilizing deep RL does present a number of significant challenges, and though these challenges do not preclude current applications of deep RL in robotics, they do limit its impact. Some of these challenges have partial or complete current solutions, whereas some do not. Although current deep RL methods are not as inefficient as often believed, provided that an appropriate algorithm is used and the hyperparameters are chosen correctly, efficiency and stability remain major challenges, and additional research on RL algorithm design should focus on further improving both. The use of simulation can further reduce challenges due to sample efficiency, though simulation alone does not solve all issues with robotic learning. Exploration can pose a major challenge in robotic RL, but we outline a variety of ways in which exploration challenges can be side-stepped in practical robotic control problems, from utilizing demonstrations to baseline hand-engineered controllers. Of course, not all exploration challenges can be overcome in this way, but ``solving'' the difficult RL exploration problem should not be a prerequisite for effective application of deep RL in robotics. Generalization presents a challenge for deep RL, but in contrast to arguments made in many prior works, we do not believe that this issue is any more pronounced than in any other machine learning field, and the availability of large and diverse data can enable RL policies to generalize in the same way as it enables generalization for supervised models. Indeed, deep RL is likely to have an advantage here -- if generalization is limited primarily by data quantity and diversity, automatically labeled robotic experience can likely be collected in much larger amounts than hand-labeled data.

Beyond the algorithmic challenges in deep RL, robotic deep RL also presents a number of challenges that are unique to the robotics setting: learning complex skills requires considerable data collection by the robots, which requires the ability to keep the robots operational with minimal human intervention. Conducting training without persistent human oversight is itself a significant engineering challenges, and requires certain best practices, as we discuss in Section~\ref{subsec:robot_at_scale}. This last challenge is tightly connected to designing persistent robots, as we desire for the robot to be an autonomous agent in the real world, there are many challenges that are often overlooked in simulated environments which we discuss in Section~\ref{subsec:robot_persistence}. As robots exist in the real world, they must also obey real-time constraints, which means that policies must be evaluated in parallel and with a limited time budget alongside the motion of the robot -- this presents challenges in the classically synchronous MDP model (Section~\ref{sec:asynchronous}). Finally, and importantly, real-world RL requires to define a reward function. Although it is common in RL research to assume that the reward function or reward signal is an external signal that is provided by the environment, in robotic learning this function must itself be programmed, or otherwise learned by the robot. As we expand the number of tasks we want our robots to accomplish via techniques such as multi-task or meta-learning discussed in Section~\ref{subsec:multi_task_and_meta}, the efforts in defining those reward functions will continue to increase. This can serve as a major barrier to deployment of RL algorithms in practice, though it can be mitigated with a variety of automatic and semi-automatic reward acquisition methods, as discussed in Section~\ref{sec:rewards}.

We believe that these challenges, though addressed in part over the past few years, offer a fruitful range of topics for future research. Addressing them will bring us closer to a future where RL can enable any robot to learn any task. This would lead to an explosive growth in the capabilities of autonomous robots -- when the capabilities of robots are limited primarily by the amount of robot time available to learn skills, rather than the amount of engineering time necessary to program them, robots will be able to acquire large skill repertoires. A suitable goal for robotic deep RL research would be to make robotic RL as natural and scalable as the learning performed by humans and animals, where any behavior can be acquired without manual scaffolding or instrumentation, provided that the task is specified precisely, is physically possible, and does not pose an unreasonable exploration challenge.


\bibliographystyle{SageH}
\bibliography{references.bib}

\end{document}

%% file: defs.tex
\newcommand{\cmt}[1]{{\iffalse\footnotesize\textcolor{red}{#1}}\fi}

\newcommand{\new}[1]{#1}
\newcommand{\old}[1]{}
\newcommand\img[1]%
  {\raisebox
    {\dimexpr-0.5\height+0.5ex}%
    {\includegraphics[width=0.95\textwidth]{#1}}
  }
\newcounter{row}

\newenvironment{imgrows}[1][\textwidth]%
  {\begin{minipage}{#1}%
   \setcounter{row}{0}%
   \stepcounter{figure}%
  }%
  {\addtocounter{figure}{-1}%
   \end{minipage}%
  }
\newcommand\imgrow
  {\par\noindent
   \refstepcounter{row}%
   \makebox[1.5em][r]{(\alph{row})}\
  }

\newcommand\literallabel[2]{%
  \label{#1}
  \expandafter\gdef\csname literallabel_#1\endcsname{#2}
  #2
}
\newcommand\literalref[1]{$\csname literallabel_#1\endcsname$}